%% file: main.tex
\documentclass[11pt]{article}

\usepackage[]{acl}

\usepackage{times}
\usepackage{latexsym}

\usepackage[T1]{fontenc}

\usepackage[utf8]{inputenc}

\usepackage{microtype}

\usepackage{amsmath}
\usepackage{enumitem}
\usepackage{float}
\usepackage{graphicx}
\usepackage{multirow}

\usepackage[linesnumbered, ruled]{algorithm2e}
\usepackage[T1]{fontenc}
\usepackage{dsfont}

\usepackage{mdframed}
\usepackage{booktabs}
\usepackage{graphicx}
\usepackage{subcaption}

\usepackage[normalem]{ulem}
\usepackage{enumitem}
\usepackage{xcolor}
\usepackage{soul}
\colorlet{soulred}{red!30}
\sethlcolor{soulred}%

\usepackage{amsmath,amsthm,amsfonts,amssymb,bm}
\usepackage{framed}
\usepackage{varioref}
\usepackage{float}
\usepackage{multirow}
\usepackage{dashrule}
\usepackage{url}
\usepackage{pifont}
\usepackage{helvet}

\newcommand{\ie}{\emph{i.e., }}
\newcommand{\eg}{\emph{e.g., }}

\input{sections/00_definitions}

\title{
FIFA: Unified Faithfulness Evaluation Framework \\for Text-to-Video and Video-to-Text Generation
}

\author{Liqiang Jing$^1$, {\bf Viet Lai}$^2$, {\bf Seunghyun Yoon}$^2$, {\bf Trung Bui
}$^2$, {\bf Xinya Du}$^1$ \\   $^1$University of Texas at Dallas, $^2$Adobe Research \\ jingliqiang6@gmail.com, xinya.du@utdallas.edu}

\begin{document}
\maketitle


\begin{abstract}
Video Multimodal Large Language Models (VideoMLLMs) have achieved remarkable progress in both Video-to-Text and Text-to-Video tasks. However, they often suffer from hallucinations, generating content that contradicts the visual input. Existing evaluation methods are limited to one task (\eg V2T) and also fail to assess hallucinations in open-ended, free-form responses. To address this gap, we propose FIFA, a unified \textbf{F}a\textbf{I}th\textbf{F}ulness ev\textbf{A}luation framework that extracts comprehensive descriptive facts, models their semantic dependencies via a Spatio-Temporal Semantic Dependency Graph, and verifies them using VideoQA models. We further introduce \corrector, a tool-based correction framework that revises hallucinated content. Extensive experiments demonstrate that FIFA aligns more closely with human judgment than existing evaluation methods, and that \corrector~ effectively improves factual consistency in both text and video generation. Our code is available at \url{https://github.com/du-nlp-lab/FIFA}.
\end{abstract}

\input{sections/1_intro}

\input{sections/2_related_work}

\input{sections/3_analysis}

\input{sections/4_method}

\input{sections/5_results}

\input{sections/6_conclusion}

\section*{Limitations}
\ours~ focuses primarily on factual precision, ensuring that each piece of information in a text is supported by the visual input. Factual recall is more challenging and an open question \cite{DBLP:conf/emnlp/MinKLLYKIZH23}.



\bibliography{custom}
\bibliographystyle{acl_natbib}

\input{sections/7_appendix}

\end{document}

%% file: sections/00_definitions.tex
\usepackage{booktabs}
\usepackage{graphicx}

\usepackage{xcolor}
\usepackage{listings}
\usepackage{fancyhdr}
\usepackage{multirow}
\usepackage{amsmath} 
\usepackage{courier} 
\usepackage{pdfpages}
\usepackage{adjustbox}
\usepackage{float}
\usepackage{lscape}

\usepackage{pifont}

\usepackage{amsmath}
\usepackage{bbm}
\usepackage{bm}

\usepackage{makecell}
\usepackage{wrapfig}
\usepackage{longtable}

\usepackage{float} 

\definecolor{codebg}{rgb}{0.95,0.95,0.95}
\definecolor{keywords}{rgb}{0,0,1}    
\definecolor{comment}{rgb}{0.5,0.5,0.5} 
\definecolor{string}{rgb}{0.58,0.0,0.82} 

\newcommand{\ours}{{FIFA}}
\newcommand{\corrector}{{Post-Correction}}

\usepackage[most]{tcolorbox}

\definecolor{assistantonecolor}{RGB}{19,118,188}
\definecolor{assistanttwocolor}{RGB}{229,91,43}

\newcommand{\assistanttwomsg}[1]{\textcolor{assistanttwocolor}{{\textbf{#1: }}}}

\usepackage{verbatim}  

\usepackage{xcolor}
\usepackage{listings}

\tcbset{
  aiboxbreakable/.style={
    width=\columnwidth, 
    top=10pt,
    colback=black!05,
    colframe=black!20,
    colbacktitle=black!50,
    enhanced,
    center,
    breakable,
    attach boxed title to top left={yshift=-0.1in,xshift=0.15in},
    boxed title style={boxrule=0pt,colframe=white,},
  }
}
\newtcolorbox{AIBoxBreak}[2][]{aiboxbreakable,title=#2,#1}

%% file: sections/1_intro.tex
\section{Introduction}

Video Multimodal Large Language Models (VideoMLLMs) \citep{video-chatgpt,videollama1} have demonstrated impressive performance across a wide range of video tasks, such as Video-to-Text (V2T) \citep{videogpt} and Text-to-Video (T2V) \citep{sora}.
Although VideoMLLMs have demonstrated remarkable performance, they are often susceptible to hallucinations, \ie the fabricated or inaccurate content generation \citep{DBLP:journals/corr/abs-2406-16338}. 
Such hallucinations pose serious risks, potentially leading to misinformation and safety concerns, and ultimately undermining the reliability of these models in real-world applications.
Despite the criticality of this issue, limited research has focused specifically on hallucination in VideoMLLMs \cite{DBLP:journals/corr/abs-2412-03735}. Existing studies mainly leveraged existing Video Question Answering (VideoQA) datasets or constructed specialized datasets for hallucination evaluation in VideoMLLMs \citep{DBLP:journals/corr/abs-2406-16338,DBLP:journals/corr/abs-2412-03735}. 

\begin{figure*}
    \centering
\includegraphics[width=\linewidth]{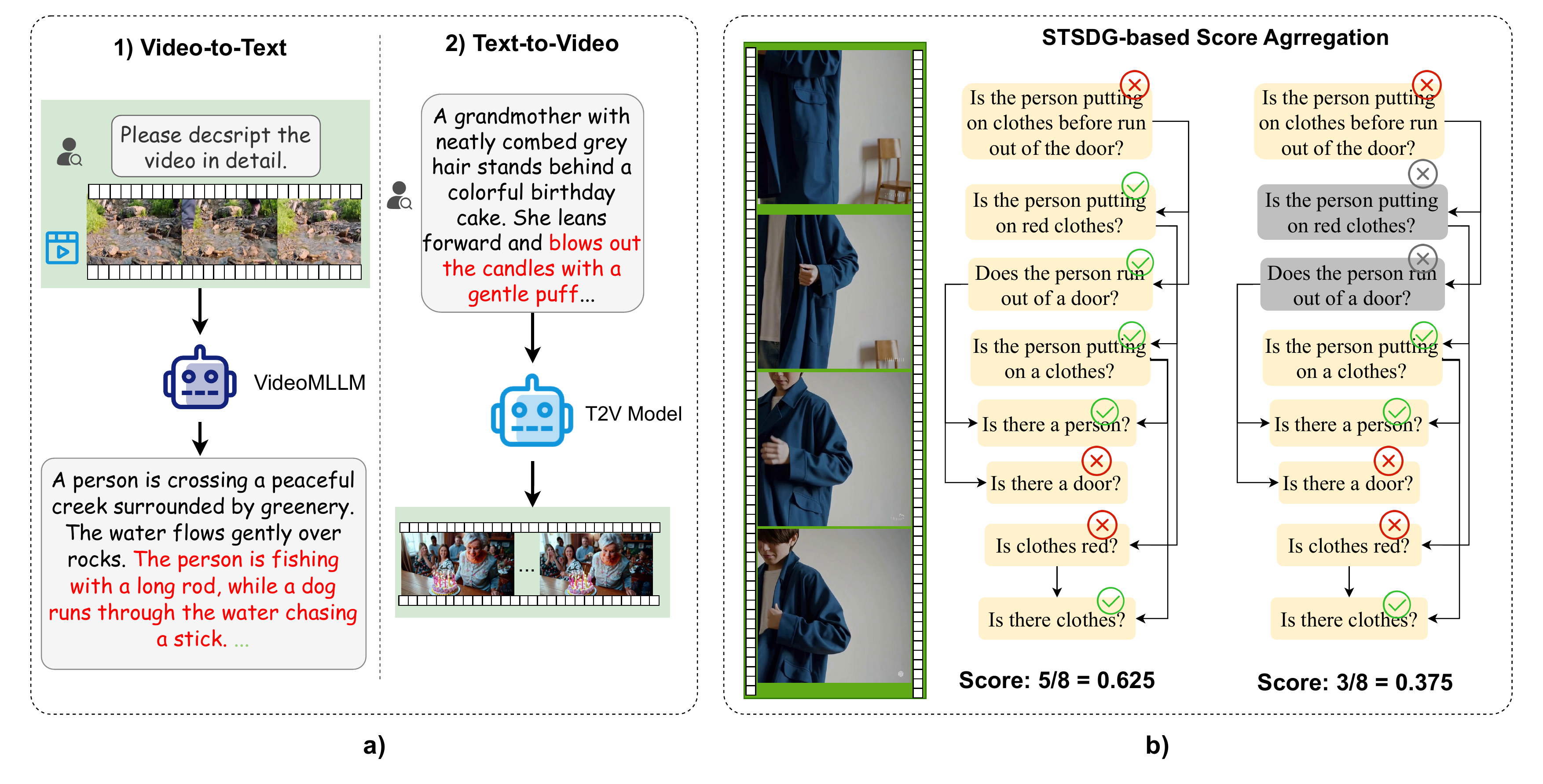}
    \caption{a) Illustration of the V2T and T2V tasks. The content in red font denotes hallucinated content. b)  Illustration of Spatio-Temporal Semantic Dependency Graph-based Score Aggregation. }
    \label{fig:task_formulation}
    \vspace{-2mm}
\end{figure*}

Although multiple works have worked on hallucination evaluation in VideoMLLMs, existing efforts are relatively isolated and face notable limitations. \textbf{First}, most approaches are restricted to simplified evaluation settings, such as binary-labeled VideoQA \citep{DBLP:journals/corr/abs-2406-16338}. As a result, they fail to address hallucinations in complex, free-form, and long-form responses to open-ended questions, scenarios that more accurately reflect real-world usage. \textbf{Second}, current research predominantly targets the V2T models \citep{DBLP:journals/corr/abs-2412-03735,DBLP:journals/corr/abs-2406-16338}, while overlooking T2V generation. 
Despite their significance to the development of general artificial intelligence, hallucinations in video generation tasks remain an underexplored issue.

To develop a unified evaluation framework for both T2V and V2T tasks involving free-form questions, 
we resort to decomposition-based evaluation methods, which first break down a response into smaller atomic information units (\ie atomic facts) and then verify each unit individually.
However, designing such a framework for VideoMLLMs is non-trivial due to the following three challenges:
\begin{itemize}[leftmargin=*]
    \item \textbf{Full Semantic Coverage}:
    On the one hand, the existing work focuses on static scenes \cite{DBLP:conf/emnlp/JingLCD24,tifa}, overlooking the hallucination in video dynamic scenes, such as temporal hallucination. On the other hand, they typically rely on atomic units, which may fail to capture the full meaning, potentially overlooking hallucinations during video-related tasks. 
    For example, for the V2T task, consider a video where {\it ``There are two people, one is wearing red clothes and the other is wearing a blue hat.'',}  The predicted video description is {\it ``There are two people; one is wearing red clothes and a blue hat.''} When decomposed into atomic information units such as {\it ``two people''}, {\it ``red clothes''}, {\it ``blue hat''}, {\it ``one person wears red clothes''}, and {\it ``one person wears a blue hat''}, each individual unit might appear faithful compared to video content. However, the predicted video description contains a hallucination: it incorrectly attributes both the red clothes and the blue hat to the same person. This inter-fact contradiction is missed in the evaluation process. There are also similar situations in the T2V task.
    \item \textbf{Dependency Between Information Units}: In the video-related task, the correctness of facts derived from text tends to depend on the others. For instance, the statement {\it ``The dog is white''} presumes that {\it ``There is a dog''} is true. If the model hallucinates the existence of a dog, then any attributes ascribed to it, such as color—are also hallucinated by implication. Without explicitly modeling these dependencies, the evaluation may produce inconsistencies. For example, if {\it ``There is a dog''} is (correctly) identified as hallucinated, yet {\it ``The dog is white''} is (incorrectly) judged as faithful, the evaluation fails to capture the inherent dependency between the two facts.
    \item \textbf{Complexity of Responses}: Unlike closed-domain tasks such as binary VideoQA \citep{DBLP:journals/corr/abs-2412-03735}, answering open-ended video-related questions often requires not only describing visual content but also providing analytical reasoning that incorporates external commonsense knowledge.  These subjective or abstract elements go beyond direct observation and can confound factuality judgments if not properly separated from descriptive content.  Failing to distinguish between analytical and descriptive content inevitably distracts the factual measurement. 
\end{itemize}

\textbf{To tackle the above challenges, we propose FIFA, a unified faithfulness metric for T2V and V2T with a Spatio-Temporal Semantic Dependency Graph.} FIFA first extracts a comprehensive set of facts from the generated text or text instruction, including both atomic facts (including temporal hallucinations) and event-level facts 
(a kind of composite fact including all associated atomic facts/information of core objects in an event)
that better captures the full semantics of the text. We instruct LLMs to extract only descriptive facts to avoid evaluation bias caused by subjective or analytical content. Subsequently, we construct a Directed Acyclic Graph (DAG), the Spatio-Temporal Semantic Dependency Graph (STSDG), by linking fact pairs that exhibit semantic dependency relationships. Next, we transform the extracted facts into questions and utilize state-of-the-art VideoQA models to answer them based on the given video content. Finally, we aggregate the verification results of all questions using the constructed STSDG to derive the overall faithfulness score. 
These dependencies ensure the consistency that if the answer to a prerequisite question is negative, all downstream questions that depend on it are skipped during evaluation, thus preventing invalid fact verification and ensuring reliable scoring.

To evaluate \ours, we conduct human annotation to assess hallucinations in both T2V and V2T tasks. We then compute the correlations between human judgments and various baseline methods. 
\ours~ yields the highest correlation with human evaluations compared to existing metrics across T2V and V2T tasks.
To further validate the key components of \ours, we construct several dedicated evaluation sets targeting different stages of the pipeline, including Fact Extraction, Fact-to-Question Generation, VideoQA, and Dependency Generation.
In addition, we introduce a unified correction framework, \corrector, which could utilize our  \corrector\ intermediate evaluation results to mitigate hallucinations in both generated video and text outputs. Extensive experiments confirm the effectiveness of our full pipeline in enhancing the factuality and reliability of generated content.

Our contributions are summarized as:
    1) To the best of our knowledge, we are the first to propose a unified evaluation metric that jointly addresses both Video-to-Text and Text-to-Video tasks.
    2) We construct a STSDG to explicitly model dependencies between a comprehensive set of facts, thereby enhancing the robustness and reliability of the evaluation process.
    3) We are the first to develop a unified correction framework, \corrector, which can identify hallucinated content and revise it to improve the factual consistency of both generated text and video.
    4) We conduct comprehensive experiments, and the results demonstrate the effectiveness of both our proposed \ours~ metric and the hallucination mitigation strategy.
    5) We created a human-annotated dataset that could facilitate future research on video-based multimodal hallucination and faithfulness evaluation.

%% file: sections/2_related_work.tex
\section{Related Work}

{\bf Video-to-Text Generation.} Video-ChatGPT \citep{video-chatgpt} applies spatial-temporal pooling to extract relevant video features, while Video-LLaMA \citep{videollama1} introduces a Video Q-Former to summarize frame-level information. Vista-LLaMA \citep{vista-llama} enhances the alignment between visual and language modalities by maintaining equal attention distances and further proposes a temporal Q-Former for temporal reasoning. LLaMA-VID \citep{llama-vid}, on the other hand, adopts a dual-token design, assigning each frame both a context and a content token, which aids in modeling long-range temporal dependencies. Despite their promising results on several benchmarks, these models still exhibit hallucinations \citep{DBLP:journals/corr/abs-2406-16338}.

{\bf Text-to-Video Generation.} 
Early studies 
such as TGANs-C \citep{tgansc} and VQ-VAE \citep{vqvae}
generate short videos with some temporal coherence.
Diffusion-based model, e.g., VDM \citep{vdm}, MagicVideo \citep{magicvideo}, PixelDance \citep{PixelDance}, and VideoCrafter2 \citep{DBLP:conf/cvpr/ChenZCXWWS24}, leverage latent diffusion and temporal attention to generate high-fidelity videos with improved temporal consistency. In parallel, autoregressive transformers \citep{vaswani2023attentionneed}, such as NUWA \citep{DBLP:conf/eccv/WuLJYFJD22}, Phenaki \citep{DBLP:conf/iclr/VillegasBKM0SCK23}, and VideoGPT \citep{videogpt}, model video sequences as discrete latent tokens, allowing better handling of temporal structure and long-context reasoning.
While these methods have greatly improved video generation quality, they often produce hallucinated content, including objects, attributes, or actions that do not faithfully reflect the input prompt. 
This hallucination issue presents a serious challenge for practical applications where semantic consistency and factual grounding are essential \citep{DBLP:journals/corr/abs-2401-07781,vbench2}.

\begin{figure*}
    \centering
    \includegraphics[width=\linewidth]{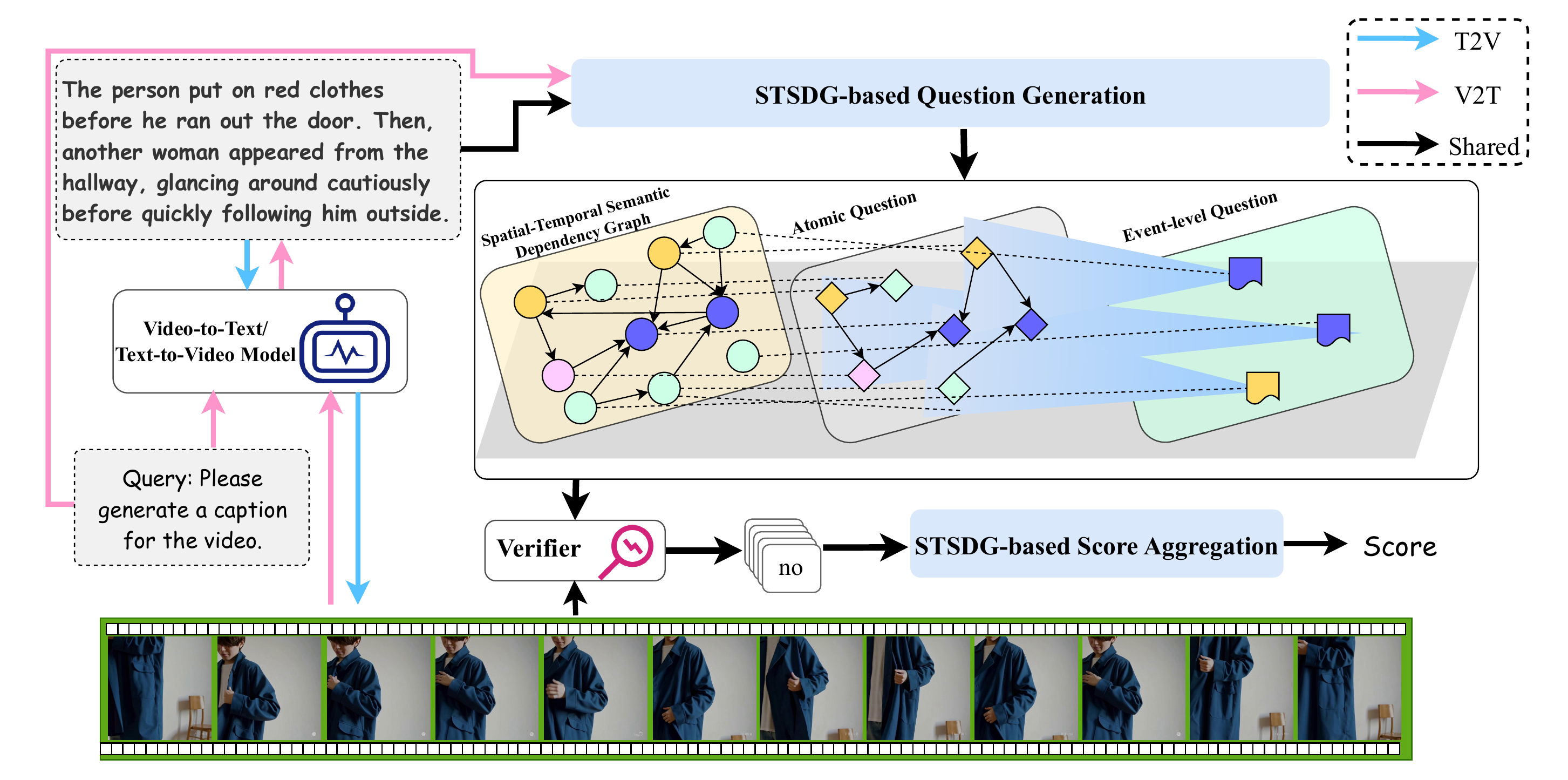}
    \caption{Illustration of our proposed \ours~ metric. The blue arrows represent the information flow for T2V, the pink arrows represent the flow for V2T, and the black arrows are shared information pathways of both tasks. }
    \label{fig:method}
    \vspace{-2mm}
\end{figure*}

{\bf MLLM Hallucination.} Hallucination
is a persistent issue in large language models (LLMs) \citep{DBLP:journals/corr/abs-2311-05232} and MLLM \citep{videollama1}.
Early studies primarily focus on hallucinations in image-related tasks \citep{DBLP:conf/emnlp/JingLCD24,tifa,dsg,DBLP:conf/iclr/LiuLLWYW24}. For example, Woodpecker \citep{DBLP:journals/chinaf/YinFZXWSSLSC24} 
refines generated responses using additional visual evidence. Similarly, Volcano \citep{DBLP:conf/naacl/LeePJS24} employs a self-refinement pipeline comprising critique, revision, and decision phases to enhance the factual accuracy of model outputs. 
Recently, the research community has investigated hallucination evaluation for video-related tasks \citep{vbench2,DBLP:conf/accv/UllahM22,DBLP:journals/corr/abs-2409-16597,DBLP:journals/corr/abs-2411-10867,DBLP:journals/corr/abs-2412-03735,DBLP:journals/corr/abs-2406-16338}. 
Different from them, we propose a unified reference-free faithfulness evaluation framework with a spatio-temporal semantic dependency graph for both V2T and T2V. We also propose a \corrector\ method to mitigate the hallucination in the generated video and text.




%% file: sections/3_analysis.tex
\section{Unified Fine-grained Faithfulness Evaluation Framework}



This section presents a unified fine-grained faithfulness evaluation metric with STSDG.
This metric evaluates the fine-grained hallucination in T2V and V2T models. Specifically, our FIFA consists of three components: STSDG-based Question Generation, Fact Verification, and STSDG-based Score Aggregation (See Figure \ref{fig:method}).

\subsection{Unified Faithfulness Evaluation Problem Formulation}
\textbf{Tasks.} Firstly, we formulated the text-to-video task and the video-to-text task. 1) \textbf{Video-to-Text.} 
Given an input video $V_t$ and a corresponding query $Q$, the video-to-text task aims to generate a response $T_t$ from a large-scale video-language model $\mathcal{M}_t$ as follows: $\mathcal{M}_t(V_t, T_p) \rightarrow T_t$.
2) \textbf{Text-to-Video.} Given an input text $T_v$, the text-to-video task aims to generate a video $V_v$ from a large-scale video generation model $\mathcal{M}_v$ as follows: $\mathcal{M}_v(T_v) \rightarrow V_v$.

\textbf{Unified Evaluation Metric.} Our goal is to develop a novel unified faithfulness metric, which necessitates the check of each video-text pair $a = (V, T)$, wherein $V$ denotes either the visual input provided to a large video-language model, or the visual output synthesized by a large video generation model.
Formally, the faithfulness score is defined as follows,
$f = \mathcal{F}(V, T,  Q)$, 
where $f$ is a scalar ranging from 0.0 to 1.0--higher values indicate greater faithfulness and fewer hallucinations in the model output. $\mathcal{F}(\cdot)$ is the faithfulness estimation, which takes video, text, and input query $(V, T, Q)$ or video and text $(V, T)$ as inputs. 
$Q=\phi$ for the text-to-video task.
Importantly, we make the proposed evaluation approach reference-free, meaning it does not rely on ground-truth annotations or human-written answers, making it broadly applicable across diverse video-based tasks.

\subsection{STSDG-based Question Generation}
We introduce how we generate various questions and the STSDG, as shown in Figure \ref{fig:DSG}.

\subsubsection{Extensive Semantic Fact Extraction}

To enable fine-grained faithfulness evaluation, we introduce an extensive semantic fact extraction module that segments the response into atomic factual units. 
Inspired by prior works \citep{DBLP:conf/emnlp/MinKLLYKIZH23}, we define an atomic fact as the smallest indivisible unit of meaning. Furthermore, in the context of T2V and V2T, we categorize atomic facts as entities, attributes, relations, or scenes. 
This granularity ensures that each piece of information can be individually assessed for accuracy without interference from unrelated content. Specifically: 
{\bf Entity} facts express the presence or absence of specific objects, including a whole entity or part of an entity(\eg door, man, and tree). 
{\bf Attribute} facts refer to object characteristics, including type, material, count, color, shape, texture, and size (\eg wooden door and red chair). 
{\bf Relation} facts describe interactions or spatial-temporal relationships between entities, including spatial relation, action, and temporal relation (\eg the man picks up the book). 
{\bf Scene} facts reflect global properties of the scene, such as lighting condition (e.g., bright lighting), overall composition, or atmosphere (\eg the atmosphere looks happy).


To evaluate the hallucination precisely, another principle is full semantic coverage: all contents of possible hallucination for the prompt, and only the contents of the prompt, should be represented by the generated questions. 
However, only these kinds of hallucinations are sometimes not enough to demonstrate real faithfulness for text-video pairs. Just as we mentioned in the example in the introduction.
Therefore, we additionally introduce another type of fact: the event-level fact.

{\bf Event-level facts} are composite facts capturing high-level semantics that cannot be expressed by a single atomic fact alone. 
An event-level fact involves multiple core objects (typically an action or relation). 
Then, all associated semantic information about these core objects, such as their attributes, states, locations, or other relations, is integrated into a single holistic fact. 
This abstraction allows for disambiguation and full interpretation of complex visual events, which would otherwise be underspecified using only atomic facts. 
Start with an atomic fact (e.g., {\it a person runs out of a door}), and enrich it by aggregating other atomic facts associated with each object in that atomic fact (e.g., {\it ``The person looks sad''} and {\it ``The door is green''}) into one comprehensive fact (e.g., {\it a sad person runs out of a green door}).
These facts are designed to cover the full meaning of a text, especially when multiple entities, relations, or temporal logic are involved, hence covering semantics when atomic-level representations fall short.

We leverage an LLM to extract all facts from the descriptive text \citep{DBLP:conf/emnlp/MinKLLYKIZH23}.
Meanwhile, we explicitly instruct the LLMs to exclude any facts that involve non-descriptive content during generation.
We construct a few-shot prompt by annotating a set of $K_1$ demonstration examples, and use them to guide the LLM in decomposing descriptive sentences into fine-grained facts. Formally, given the text $T$ for the T2V task/the text and query $(T, Q)$ for the V2T task, we obtain fact groups:
\begin{equation}
	G=\begin{cases}
	\text{LLM}(P_{t2v}, T), task=t2v,\\
	\text{LLM}(P_{v2t}, T, Q), task=v2t,
	\end{cases}
    \label{eq1}
\end{equation}
where $G = \{g^1, \cdots, g^{n}\}$ denotes the set of $n$ generated facts. $P_{t2v}$ and $P_{v2t}$ are in-context instruction of fact extraction for the T2V task and V2T task, respectively (See Appendix \ref{appendix:prompts} for prompt template).


\subsubsection{STSDG Construction}
\begin{figure}
    \centering
    \includegraphics[width=1\linewidth]{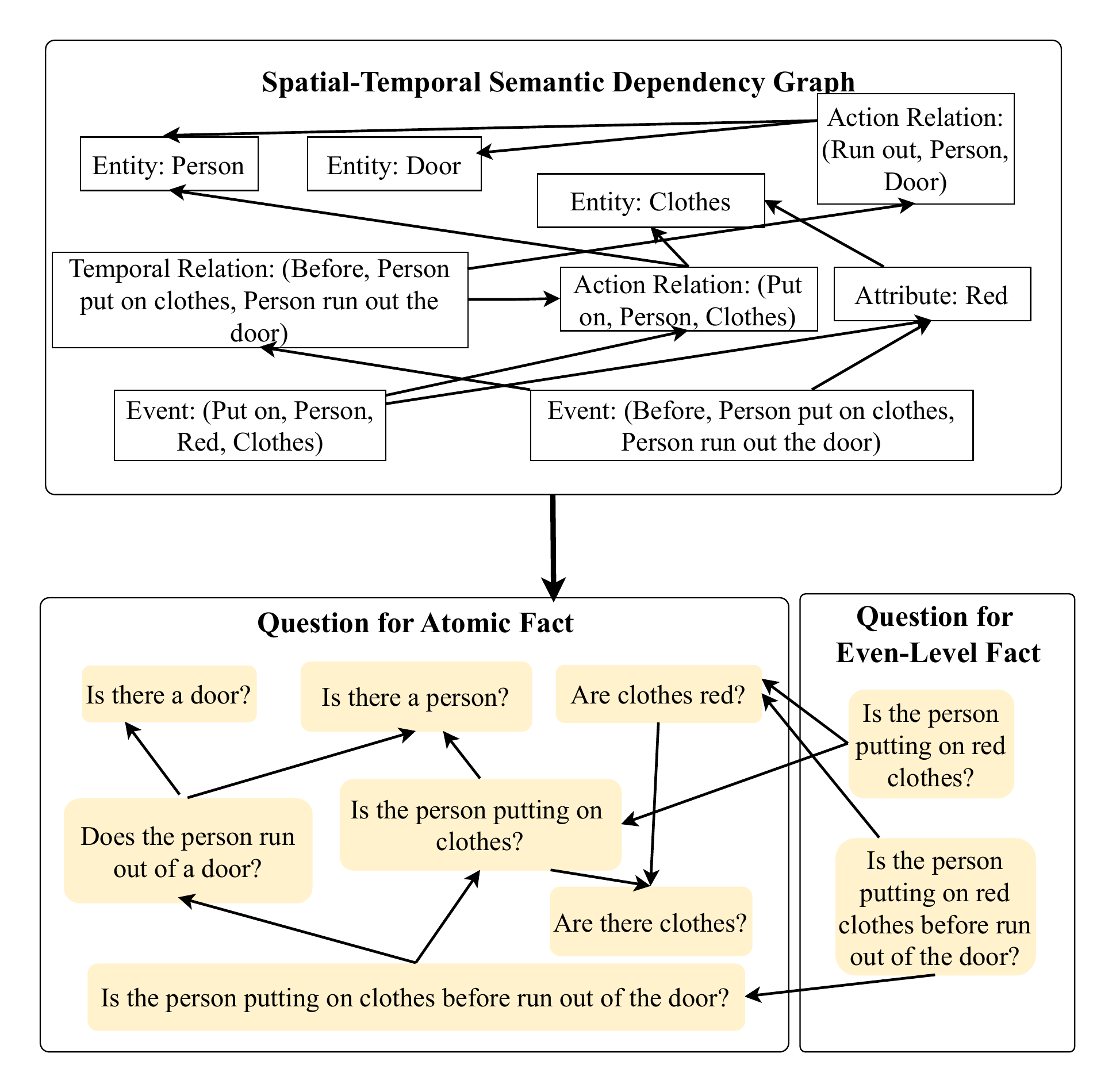}
    \caption{Illustration of STSDG-based Question Generation and  STSDG-based Score Aggregation. The generation process is for the text ``The person put on red  clothes before he ran out the door.''.}
    \label{fig:DSG}
    \vspace{-2mm}
\end{figure}

To verify the faithfulness of all facts, we further convert them into a yes-or-no question in natural language format with LLM as 
$\{q_1, \cdots, q_n\} = LLM(P_q, G, T, Q)$, 
where $Q=\phi$ for the text-to-video task. $P_q$ is the prompt and is shown in Appendix \ref{appendix:prompts}. $q_i$ is the generated question for the $i$-th fact.
As we mentioned before, there are semantic relationships between different facts/questions, which could improve the reliability of our metric. Therefore, in this component, we construct an STSDG (see Figure \ref{fig:DSG}) to model dependent relationships between questions.

Briefly sketched, the STSDG is a set of Text-Video alignment validation questions structured in a directed scene graph, produced from the text as the ground truth. 
In particular, we deem the generated question as nodes in the graph, denoted as $\mathcal{Q} = \{q_1,\cdots, q_n\}$. 
Next, we generate the edges for the nodes.
Specifically, similar to the last step, we also implemented this stage by an LLM given task-specific in-context examples: we prompt an LLM with a preamble (with input and output sampled from manual annotations with fixed seeds) to elicit annotations of the same format for new inputs. 
The details on the preamble engineering is
in Appendix \ref{appendix:prompts}. Specifically, we obtain semantic dependency edges between questions as an adjacency matrix $\mathbf{E} \in \mathbbm{R}^{n\times n}$,
\begin{equation}
{E}_{ij}=\left\{
\begin{aligned}
1 & ,\quad if\ \mathcal{S}(q_i,q_j),\\
0 & ,\quad otherwise,
\end{aligned}   
\right.
\end{equation}
where $ i,j \in [1,n]$, and $\mathcal{S}(q_i,q_j)$ is True when the semantics of the question $q_i$ is depend on the question $q_j$. Notably, $\mathbf{E}$ is the adjacency matrix of a directed acyclic graph, which means $\mathbf{E}_{ij}==\mathbf{E}_{ji}$ does not necessarily hold true. 

\subsection{Fact Verification}
Based on the video content, we verify the faithfulness of each fact by answering the generated question with a VideoQA model as 
follows, 
\begin{equation}    
    A = \{a_1, \cdots, a_n\} = \operatorname{VideoQA}(V, {q_1, \cdots, q_n}), \label{eq4}
\end{equation}
where $\operatorname{VideoQA}(\cdot)$ is a video question answering model. $\{q_1, \cdots, q_n\}$ is the question set and $q_i$ is the $i$-th question corresponding $i$-th fact. $A$ is the corresponding answer for the question set.
The reason why we use VideoQA Models to verify the consistency between fact and video, even if the VideoQA may also introduce hallucination: Our method converts the AI labeling task into a discriminative task that usually generates a short response (``yes'' or ``no''), and this kind of task tends to generate low hallucination \citep{DBLP:conf/emnlp/MinKLLYKIZH23,DBLP:conf/emnlp/JingLCD24}.
 
\subsection{STSDG-based Score Aggregation}

Finally, we calculate the faithfulness score \ours~ for all the derived facts. In particular, we first convert answers $A=\{a_1, \cdots, a_n\}$ into scores $S=\{s_1, \cdots, s_n\}$. Thereafter, we utilize the semantic dependency relation to derive the refined scores to improve the reliability of the fact verification:
\begin{align}
    \hat{s}_i = \mathbb{I} (a_i=yes)  \prod_{j \text{ s.t. } E_{ij} = 1} s_j,
    \label{eq5}
\end{align}
where  $\mathbb{I}(\cdot)$ is the indicator function, and the value of $\mathbb{I} (a_i=yes)$ is 1 when $a_i$ is ``yes''. $i, j \in [1, n]$ and $i \neq j$.
Then the final faithfulness score  $\hat{f}$ is the average of all refined scores:
$\hat{f} = 1/n {\sum_{i=1}^{n}  {\hat{s}_i} }$.

%% file: sections/4_method.tex
\section{Meta Evaluation for FIFA}

\subsection{Evaluation Setup}

We evaluate four widely-used models: two T2V models: CogVideoX \citep{cogvideo} and HunyuanVideo \citep{hunyuanvideo}, and two V2T models: Video-LLaVA \citep{DBLP:conf/emnlp/LinYZCNJ024} and Video-LLaMA \citep{videollama1}. For each task, we have 60 evaluation samples, yielding a total of 120 annotated samples of hallucination across T2V and V2T. More details are in Appendix \ref{appendix:setup_meta}



To evaluate the superiority of our proposed metric \ours, we compare it with several T2V and V2T evaluation metrics.
For V2T metrics, we compare \ours~ with 1) reference-based: BLEU-4~\citep{bleu}, ROUGE-L~\citep{ROUGE}, METEOR~\citep{meteor}, BERT-Score \citep{bert-score}, and COAHA \citep{DBLP:conf/accv/UllahM22}; and 2) reference-free: CLIP-Score~\citep{clipscore}. 
For T2V metrics, it is harder to collect ground-truth compared with the V2T task.
Hence, we only select reference-free metrics for comparison. We select CLIP-Score, XCLIP-Score \citep{DBLP:conf/eccv/NiPCZMFXL22}, BLIP-BLEU \citep{DBLP:conf/cvpr/LiuC0WZCLZCS24}, mPLUG-BLEU \citep{DBLP:conf/cvpr/LiuC0WZCLZCS24} and FAST-VQA \citep{DBLP:conf/eccv/WuCHLWSYL22}
as baselines.

To quantify the human evaluation of faithfulness, we employ the 1-5 Likert Scale~\citep{likert1932technique} to score the faithfulness of the text-video pair on a tangible scale, ranging from 1 (worst) to 5 (best). 
The details about the annotation process are given in the Appendix \ref{appendix:UI}. 
Table~\ref{tab:correlation} delineates the correlation between various evaluation metrics and human judgment regarding the faithfulness of  T2V and V2T. The result shows that our evaluation framework consistently achieves a significant improvement across T2V and V2T. We add more ablation studies in Appendix \ref{appendix:ablation} and detailed benchmark results in Appendix \ref{appendix:types}.

\begin{table}[]
    \centering
        \caption{Correlation between each evaluation metric and human judgment on V2T and T2V faithfulness evaluation, measured by Pearson's $r$, Spearman’s $\rho$, and Kendall’s $\tau$.
        The p-value of the significance test between our result and the baseline result is less than 0.01. }
            \resizebox{\linewidth}{!}{

    \begin{tabular}{c|cl|ccc}
    \toprule
        \bf Task & \bf Type & \bf Metrics & \bf Pearson's $r$ & \bf Kendall's $\tau$ &
    \bf Spearman's $\rho$ \\ \midrule
    \multirow{7}{*}{\bf V2T} & \multirow{5}{*}{\bf Reference-based}&
     
    BLEU-4      & 41.12  & 35.92  & 45.39 \\
    & & ROUGE-L      & 29.55  & 22.83  & 29.31 \\
    & & METEOR        & 45.74  & 35.95  & 46.10 \\
    & & BERT-Score & 43.77& 36.85 & 50.11\\ 
    & & COAHA          & -38.15 & -11.41 & -13.70 \\
    \cmidrule(lr){2-6}
    
     & \multirow{2}{*}{\bf Reference-free}&      CLIP-Score &4.58&-1.20&-1.01 \\ 
     & & \bf \ours    &\textbf{58.20}  & \textbf{53.20}  & \textbf{62.96} \\
             \midrule
     \multirow{6}{*}{\bf T2V} & \multirow{6}{*}{\bf Reference-free}& CLIP-Score  & 30.22 & 3.42&5.31\\ 
     & & XCLIP-Score  & 24.96 & 20.63 & 29.39\\ 
     & & BLIP-BLEU  & 57.67   & 43.61 & 60.90\\ 
     & & mPLUG-BLEU & -26.39 & -30.07 & -22.70\\ 
     & & FAST-VQA   & 7.65 &4.79&5.68\\
     & & \bf \ours & \textbf{67.92} & \textbf{64.25} & \textbf{77.50} \\
    \bottomrule
    \end{tabular}}
    \label{tab:correlation}
    \vspace{-4mm}
\end{table}

\subsection{STSDG-based Generation} 
In this section, we evaluate every key stage in Spatial-Temporal Semantic Dependency Graph Construction. 
We use the human evaluation 
to verify the reliability in each intermediate stage.

\textbf{Are the generated questions reliable?}
The first stage of our evaluation framework is to extract all facts and then transform them into a question format. 
Therefore, it is very important to get high-quality questions. 
To evaluate the quality of the generated questions, we define the metrics precision and recall. For each text, we employ annotators to write the corresponding facts, denoted as $C=\{c_1,\cdots,c_{n_c}\}$. Follow the definition of the last section, the generated questions are denoted as $U = \{q_1, \cdots, q_n\}$. Based on the generated questions and annotated facts, we define $\sum m_{t,q}/|Q|$ as precision and $\sum m_{t,q}/|T|$ as recall. $|Q|$ and  $|T|$ are the total number of questions and facts, respectively. $m_{t,q}=1$ if $t$ matches $q$, otherwise, it is $0$. We show the experimental results in Table \ref{tab:human_eval}. Overall, the generated questions are close to perfect in matching the source semantic fact. 
Furthermore, we compute the consistency between 3 annotators and found Fleiss' Kappa is 0.84, which indicates an almost perfect agreement between annotators.



\begin{table}
    \centering
        \caption{
    Human evaluation results of generated questions, converting facts into questions, and validity of generated dependency for T2V and V2T tasks.
    }
    \resizebox{\linewidth}{!}{
    \begin{tabular}{l|cc|c|c}
            \toprule 
            \multirow{2}{*}{\bf Task }    & \multicolumn{2}{c|}{\bf Question Generation} & \bf Fact Conversion & \bf Dependency \\ 
                    & Precision & Recall & Accuracy & Valid Ratio \\
            \midrule
            T2V     & 98.71     & 99.22 & 99.06 & 99.06 \\    
            V2T     & 95.11     & 95.22 & 99.03 & 99.03 \\
            \midrule
            All     & 96.31     & 96.55 & 99.04 & 99.04 \\
             
            \bottomrule
        \end{tabular}
    }
    \vspace{-2mm}
        \label{tab:human_eval}
\end{table}


\textbf{Can the tuple be transferred into independent questions correctly?} 
To evaluate the performance of the conversion of extracted facts into corresponding questions, we further conduct an analysis using accuracy as the evaluation metric. The results are presented in Table~\ref{tab:human_eval}.
Overall, the accuracy of converting facts into questions are close to perfect (99.88\% for T2V and 99.26\% for V2T). 
Furthermore, we compute the consistency between 3 annotators and found the Fleiss' Kappa is 0.91, which indicates an almost perfect agreement between annotators.

\textbf{Are the generated dependencies between questions valid?}
To enhance the reliability of fact verification, our method (\ours) introduces directed dependency edges between questions. Specifically, if question $q_i$ depends on question $q_j$, then $q_i$ is considered a valid VideoQA query only if the answer to the dependent question $q_j$ is positive 
(e.g.,{\it ``is the dog white?''} is only valid if the answer to {\it ``is there a dog?''} is positive). 
 To evaluate the effectiveness of the LLM in generating such dependencies, we ask human annotators to make binary judgments for each question–dependent-question pair. We show the human evaluation results in Table \ref{tab:human_eval}. Overall, the valid ratio of dependency generation are close to perfect (99.06\% for T2V and 99.03\% for V2T). 


\subsection{Performance on Fact Verification}
As the verifier in our evaluation framework, the performance of VideoQA models plays a critical role. To assess their effectiveness, we evaluate several state-of-the-art VideoQA models, including InternVL-2.5-8b \citep{internvl25}, Video-LLaMA3-7b \citep{videollama3}, Video-LLaVA-7b \citep{DBLP:conf/emnlp/LinYZCNJ024}, Qwen2.5-VL-7b/32b/72b\citep{Qwen2.5-VL}.  Specifically, we collect 555 questions from the T2V evaluation set and 714 questions from the V2T evaluation set, each paired with its corresponding video. Every question is independently annotated by three annotators, and the final label is determined using majority voting. 
The performance of all evaluated VideoQA models is reported in Table~\ref{tab:eval_vqa}. Overall, Qwen2.5-VL-72b achieves the best performance on the T2V and V2T tasks.



\begin{table}
    \centering
        \caption{
    Human evaluation for fact verification.
    }
    \resizebox{\linewidth}{!}{
        \begin{tabular}{l|cc|cc}
        \toprule
         \bf Model & \bf T2V Accuracy  & \bf V2T Accuracy & \bf Average  \\ 
         \midrule
         InternVL-2.5-8b    & 73.86 & 68.21 & 71.56\\ 
         Video-LLaVA        & 75.47 & 76.75 & 76.19\\ 
         Video-LLaMA3       & 79.46 & 79.55 & 79.51\\ 
         Qwen2.5-VL-7b      & 73.69 & 73.25 & 73.44\\ 
         Qwen2.5-VL-32b     & 77.11 & 75.73 & 76.33\\ 
         Qwen2.5-VL-72b     & 80.00 & 80.11 & 80.06\\ 
         \bottomrule
    \end{tabular}
    }

    \label{tab:eval_vqa}
\end{table}




%% file: sections/5_results.tex

\begin{figure*}
    \centering
    \includegraphics[width=\linewidth]{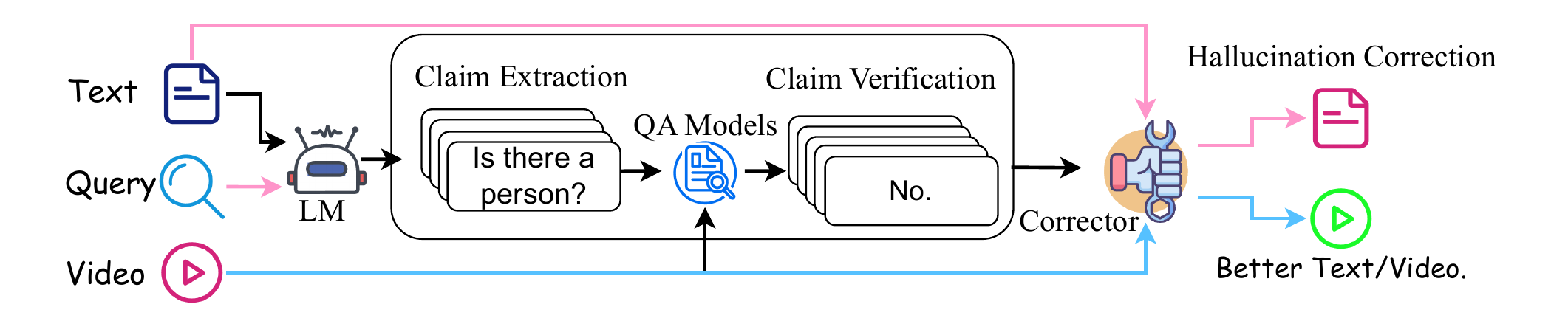}
    \caption{The proposed \corrector~ method consists of three key stages: Claim Extraction, Claim Verification, and Hallucination Correction. 
    The Corrector takes claim-answer pairs with video for T2V, and with text for V2T.
    }
    \label{fig:corrector}
\end{figure*}


\section{Post-Correction}

{\bf Method.}
Our initial experiments show various hallucinations in the T2V and V2T models. Therefore, we devise a post-correction method to alleviate these issues. 
In particular, our goal is to identify and rectify hallucinations in texts in T2V and V2T tasks. A central challenge lies in detecting hallucinated content and identifying factual information that can serve as the basis for correction. To address this, we utilize the intermediate evaluation result of our \ours~ and divide the entire process into three subtasks: key claim extraction, claim verification, and hallucination correction. An overview of our framework is shown in Figure \ref{fig:corrector}.
1)
{\bf Claim Extraction.} Since the text usually consists of multiple claims, such as objects, attributions, and relations, we follow Eq.~\ref{eq1} to extract facts from the text.
  2)   
{\bf Claim Verification.} Then, we ask a series of questions around them to make the hallucination diagnosis following operations in Eq.~\ref{eq4}. 
    For all questions, we apply a VideoQA model to answer the questions conditioned on the video.  The first two stages are the intermediate process in our \ours.
    3) 
{\bf Hallucination Correction.}
    For the \textbf{V2T} task, an LLM corrects hallucinated content in the generated textual responses. 
    Specifically, we aggregate the QA pairs into a structured prompt and instruct the LLM to generate a refined version of the response with hallucinations corrected. 
    For the \textbf{T2V} task, a video editing model is employed to revise hallucinated visual content in generated videos. 
    In particular, we first use an LLM to generate editing instructions based on the input prompt and corresponding QA pairs. 
    For example, given the input prompt {\it ``a green door''}, and QA pairs: {\it ``Is there a door? → Yes''} and {\it ``Is the door green? → No''}, the generated instruction might be {\it ``change the door to green.''} 
    The original generated video, along with this editing instruction, is then passed to a video editing model to produce a refined video.

{\bf Experiments.}
We construct evaluation sets for both  T2V and V2T tasks. For the V2T task, we sample 100 videos from the MSR-VTT dataset to perform the captioning task. For the T2V task, due to the slow generation speed of video generation models and video editing models, we adopt 30 prompts from the meta-evaluation benchmark for our experiments. We use Qwen2.5-VL-72b as the VideoQA model and TokenFlow as the video editing model in our Post-Correction method.
Table \ref{tab:correction} 
shows the performance of all the baselines without and with our correction method. For the T2V task, we found that our \ours~can improve the performance of all baselines across COAHA and FIFA metrics. 
For the V2T task, our method can also improve the FIFA and reduce hallucinations in the generated video, which demonstrates the effectiveness of our \corrector~method.
In addition, we show more benchmark results in Appendix \ref{appendix:morebench}.

\begin{table}
    \centering
    \resizebox{\linewidth}{!}{
        \begin{tabular}{c|l|cc|cc} \toprule
         \multirow{2}{*}{\bf Task} & \multirow{2}{*}{\bf Model} &  \multicolumn{2}{c|}{\bf COAHA $\downarrow$} & \multicolumn{2}{c}{\bf \ours $\uparrow$}\\
         & & w/o & w/ & w/o & w/ \\
         \midrule
         \multirow{4}{*}{V2T} &  Video-LLaVA & 52.45 & \textbf{47.23} &  63.43 & \textbf{66.08} \\
         & Video-LLaMA   & 53.34 & \textbf{45.86} & 60.46 & \textbf{65.54}\\ 
         & Video-LLaMA2  & 37.65 & \textbf{25.93} & 64.49 & \textbf{69.82}\\ 
         & Video-LLaMA3  & 63.25 & \textbf{51.27} & 65.28 & \textbf{70.41}\\              
         \midrule
         T2V & CogVideoX  & - & - & 54.53 & \textbf{60.70}\\ 
             \bottomrule
    \end{tabular}
    }
    \caption{
    Results on the V2T and T2V tasks. w/ and w/o denote whether the generated content is or is not corrected by our \corrector~ method.
    }
\label{tab:correction}
\vspace{-5mm}
\end{table}

%% file: sections/6_conclusion.tex
\section{Conclusion}
In this work, we propose \ours, a unified and reference-free faithfulness evaluation framework for both V2T and T2V tasks. 
\ours~ introduces a comprehensive fact extraction strategy and constructs an STSDG to model inter-fact relationships. These facts are then converted into questions and verified using powerful VideoQA models, with dependencies guiding the final score aggregation. Our method achieves the highest correlation with human judgments compared to existing baselines. In addition, we propose a unified correction pipeline, \corrector, to mitigate hallucinations in both generated videos and texts. 

%% file: sections/7_appendix.tex
\clearpage
\newpage
\appendix

\section{More Ablation Study}
\label{appendix:ablation}
To explore the roles of different components in our proposed evaluation framework, we compared \ours~ with the following derivations. 
1) w/o-Dependency. To explore the effect of the generated semantic dependency relation, we removed the STSDG in our evaluation framework. Specifically, we remove Equation \ref{eq5} 
from our \ours. 
2) w/-Qwen2.5-VL-7b and 3) w/-Qwen2.5-VL-32b. To verify the importance of our selected VideoQA model, we replace it with  Qwen2.5-VL-7b and Qwen2.5-VL-32b, respectively. Table \ref{tab:ablation_study} summarizes the performance of \ours~ with its derivations. From the results, we observe that: 1) Our \ours~ surpasses w/o-Dependency, demonstrating the importance of introducing semantic dependency relationships between facts/questions. 2) w/-Qwen2.5-VL-7b and w/-Qwen2.5-VL-32b perform worse than our \ours, which demonstrates the correctness of our choice of the current VideoQA model. 3) By comparing the VideoQA accuracy in Table \ref{tab:eval_vqa} and Table \ref{tab:ablation_study}, we observe that models with higher VideoQA accuracy tend to achieve better correlation performance. This suggests that improving the accuracy of the VideoQA verifier is crucial for enhancing the overall correlation between model outputs and human judgments.

\begin{table}[h]

    \centering
        \caption{Experiment results of ablation study.}
            \resizebox{\linewidth}{!}{
    \begin{tabular}{l|ccc}
    \toprule
         \bf Method& \bf Pearson's $r$& \bf Kendall's $\tau$ &
\bf Spearman's $\rho$ \\ \midrule
  \ours & \textbf{63.06} & \textbf{58.73} & \textbf{70.23} \\ \midrule
      w/o-Dependency  & 58.53 & 52.77& 66.56\\ 
      w/-Qwen2.5-VL-7b  & 56.25  & 45.66&58.06\\ 
      w/-Qwen2.5-VL-32b & 46.46 & 44.89 & 54.69 \\ 
         \bottomrule
    \end{tabular}}
    \label{tab:ablation_study}
\end{table}

\section{Experimental Setups for Meta-Evaluation}
\label{appendix:setup_meta}
\textbf{V2T Data.} We sampled videos from the validation set of the widely-used video captioning dataset MSR-VTT \citep{DBLP:conf/cvpr/XuMYR16} for human evaluation. 
To enrich the diversity of question types in our dataset, we designed different types of queries for evaluation. Specifically, we selected 10 videos for the captioning task, using the query {\it ``Please generate a brief for the video''} with the ground-truth captions from MSR-VTT serving as the reference answers.

For the remaining two tasks, i.e., detailed description and complex question answering, we sampled 10 different videos for each task and used GPT-4o to generate corresponding prompt-answer pairs, following LLaVA \citep{llava}.

\textbf{T2V Data.} We selected 20 captions from the validation set of MSR-VTT as inputs for the T2V task. However, these captions are typically short and contain limited semantic elements, such as objects, attributes, and temporal relationships.
To address this limitation, we further sampled an additional set of 10 captions and employed GPT-4o to generate richer and more informative prompts, aiming to better evaluate the model’s ability to handle complex and detailed textual inputs.

\section{Comparison with Existing Evaluation Metrics}
To evaluate the superiority of our proposed metric \ours, we compare it with several T2V and V2T evaluation metrics.
For V2T metrics, we compare \ours with 1) reference-based: BLEU-\{1/2/3/4\}~\citep{bleu}, ROUGE-\{1/2/L\}~\citep{ROUGE}, METEOR~\citep{meteor}, BERT-Score \citep{bert-score}, and COAHA \citep{DBLP:conf/accv/UllahM22}; and 2) reference-free: CLIP-Score~\citep{clipscore}. 
For T2V metrics, it is harder to collect ground-truth compared with the V2T task.
Hence, we only select reference-free metrics for comparison. Specifically, we select CLIP-Score, XCLIP-Score \citep{DBLP:conf/eccv/NiPCZMFXL22}, BLIP-BLEU \citep{DBLP:conf/cvpr/LiuC0WZCLZCS24}, mPLUG-BLEU \citep{DBLP:conf/cvpr/LiuC0WZCLZCS24} and FAST-VQA \citep{DBLP:conf/eccv/WuCHLWSYL22}
as baselines.

For all evaluation tasks, we employed three annotators to independently annotate each sample to ensure the reliability and consistency of the annotations. All GPT-4o outputs used in our experiments were generated with the model version \texttt{gpt-4o-2024-08-06}. We use Qwen2.5-VL-72b \citep{Qwen2.5-VL} as our videoQA model and use GPT-4o as the LLM in our evaluation framework.
Details of our annotation interface are provided in Appendix~\ref{appendix:UI}.

\section{Human Evaluation}
We employ 3 workers for annotation  via Amazon Mechanical Turk \footnote{\url{https://www.mturk.com/}.}. Every worker is a native English speaker. They are paid 15-20 USD per hour. Every worker went through a qualification test of 2 hours and was tested to be highly qualified. 

\begin{table*}[h]
\centering
\begin{tabular}{l|cccccccc}
\toprule
\textbf{T2V Model} & Entity & Attribute & Spatial &   Temporal& Action & Event  \\ \midrule
CogVideox& 84.07 & 77.42 & 72.22 & 60.00 & 71.70 & 63.46  \\
HunyuanVideo& 86.49& 76.67&66.67& 20.00&54.72&52.94 \\ \midrule
\midrule 
\textbf{V2T Model} & Entity & Attribute & Spatial &   Temporal& Action & Event  \\ \midrule
Video-LLaMA& 88.08 &73.53   &  71.43 & 68.00& 73.81 &58.97  \\ \midrule
Video-LLaVA& 90.35 & 90.48 & 75.86 & 56.61 & 67.65 &  57.14  \\
\bottomrule
\end{tabular}
\caption{Comparison of T2V models V2T models.}
\label{tab:accuracy_horizontal_merged}
\end{table*}

\section{Prompts}
\label{appendix:prompts}

\begin{AIBoxBreak}{Fact Extraction Prompt}
\assistanttwomsg{Prompt}
Task: given input prompts, describe each scene with skill-specific tuples. Do not
generate the same tuples again. Do not generate tuples that are not explicitly described in the prompts.

output format: id | tuple

\$\{In-context Examples\}\$
\end{AIBoxBreak}

\begin{AIBoxBreak}{Question Generation Prompt}
\assistanttwomsg{Prompt}
Task: given input prompts and skill-specific tuples, re-write tuple each in natural
language question.

output format: id | question

\$\{In-context Examples\}\$
\end{AIBoxBreak}

\begin{AIBoxBreak}{Dependency Generation Prompt}
\assistanttwomsg{Prompt}
Task: given input prompts and tuples, describe the parent tuples of each tuple.

output format: id | dependencies.

\$\{In-context Examples\}\$
\end{AIBoxBreak}

We show the concert in-context examples in Section \ref{appendix:ICL}.

\section{Experimental Details}

We run all experiments on a server with 4 $\times$ A100 GPUs.



\section{Fine-grained Benchmark for V2T and T2V}
\label{appendix:types}

Table~\ref{tab:accuracy_horizontal_merged} presents a comparison of T2V and V2T models across different fact categories in our human evaluation. We observe that entity and attribute categories achieve relatively high FIFA scores across all models, indicating that hallucinations related to objects and their properties are less frequent in video-related tasks. In contrast, action and relation categories (particularly spatial and temporal relations) tend to have lower scores, suggesting these are the main sources of hallucination. Notably, the temporal category shows the lowest accuracy in T2V settings, highlighting the importance of modeling temporal hallucinations explicitly. Additionally, the low scores in the event-level facts underscore the necessity of incorporating composite, high-level semantic facts to better capture and evaluate complex visual events.


\section{Hallucination Mitigation on More Benchmarks}
\label{appendix:morebench}
In addition to the caption task, we also conduct experiment on VideoHallucer \cite{DBLP:journals/corr/abs-2406-16338}, which is a binary-QA question answering benchmark. We show the results in Table \ref{appendix:tab:correction}. From this table, we found that our method could mitigate hallucination for all baselines. In addition, we observe a positive correlation between the performance of our method and the baseline models. In general, the stronger the baseline, the greater the improvement achieved by our approach.

\begin{table}
    \centering
    \resizebox{0.65\linewidth}{!}{
        \begin{tabular}{l|cc} \toprule
         \multirow{2}{*}{\bf Model} &  \multicolumn{2}{c}{\bf Accuracy $\uparrow$} \\
          & w/o & w/ \\
         \midrule
           Video-LLaVA & 55.75 & \textbf{56.68}  \\
          Video-LLaMA   & 53.75 & \textbf{62.25} \\ 
          Video-LLaMA2  & 56.25 & \textbf{61.75}\\ 
          Video-LLaMA3  & 65.42 & \textbf{67.02} \\              

             \bottomrule
    \end{tabular}
    }
    \caption{
    Results on the VideoHallucer benchmark, a V2T hallucination evaluation task.
    }
\label{appendix:tab:correction}
\vspace{-5mm}
\end{table}

\input{sections/tables/ICL-fact-extraction-T2V}

\section{Annotation Details}
\label{appendix:UI}

\begin{figure*}
    \centering
    \includegraphics[width=\linewidth]{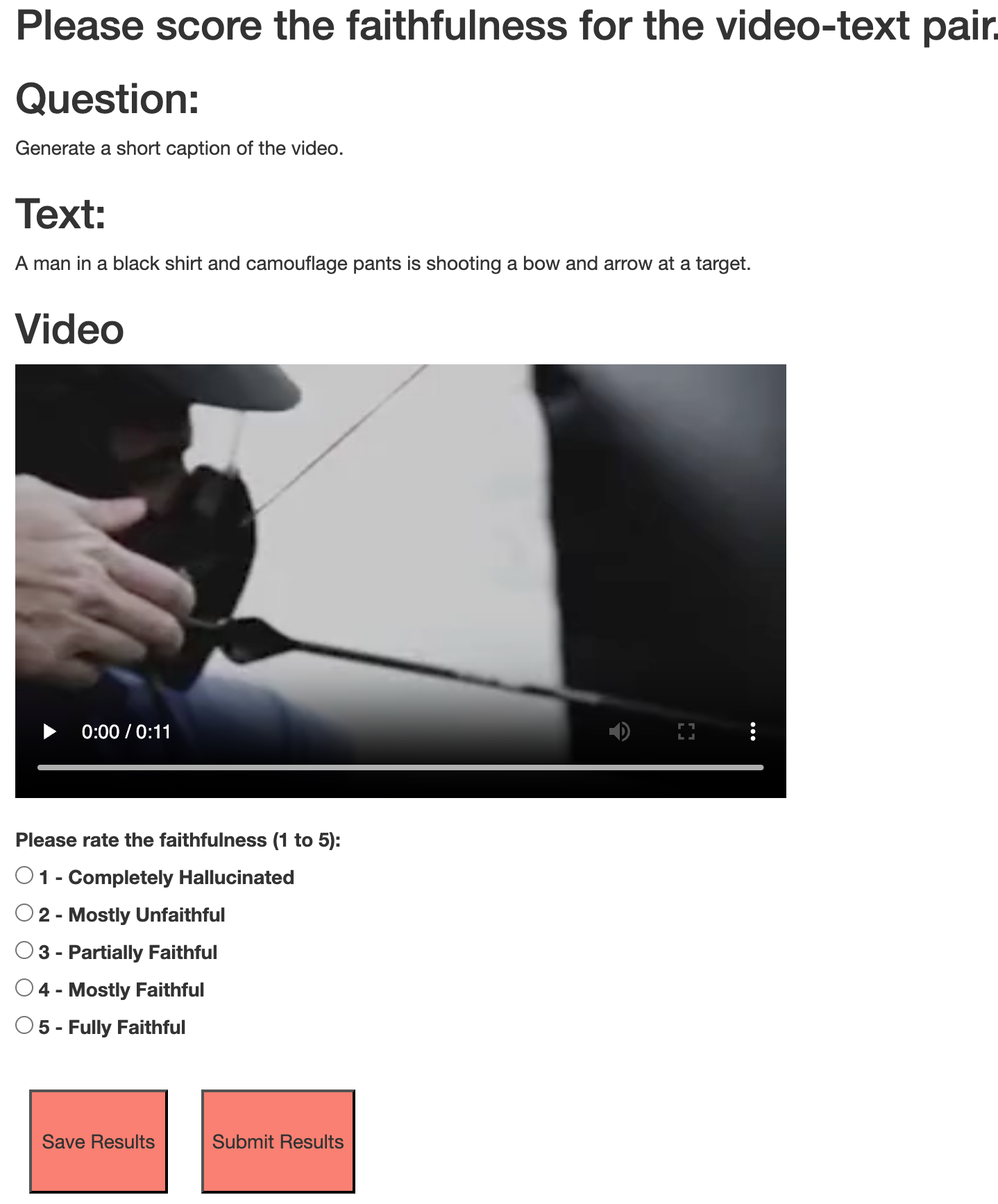}
    \caption{UI for faithfulness evaluation of human annotation.}
    \label{fig:t1_ui}
\end{figure*}

\begin{figure*}
    \centering
    \includegraphics[width=\linewidth]{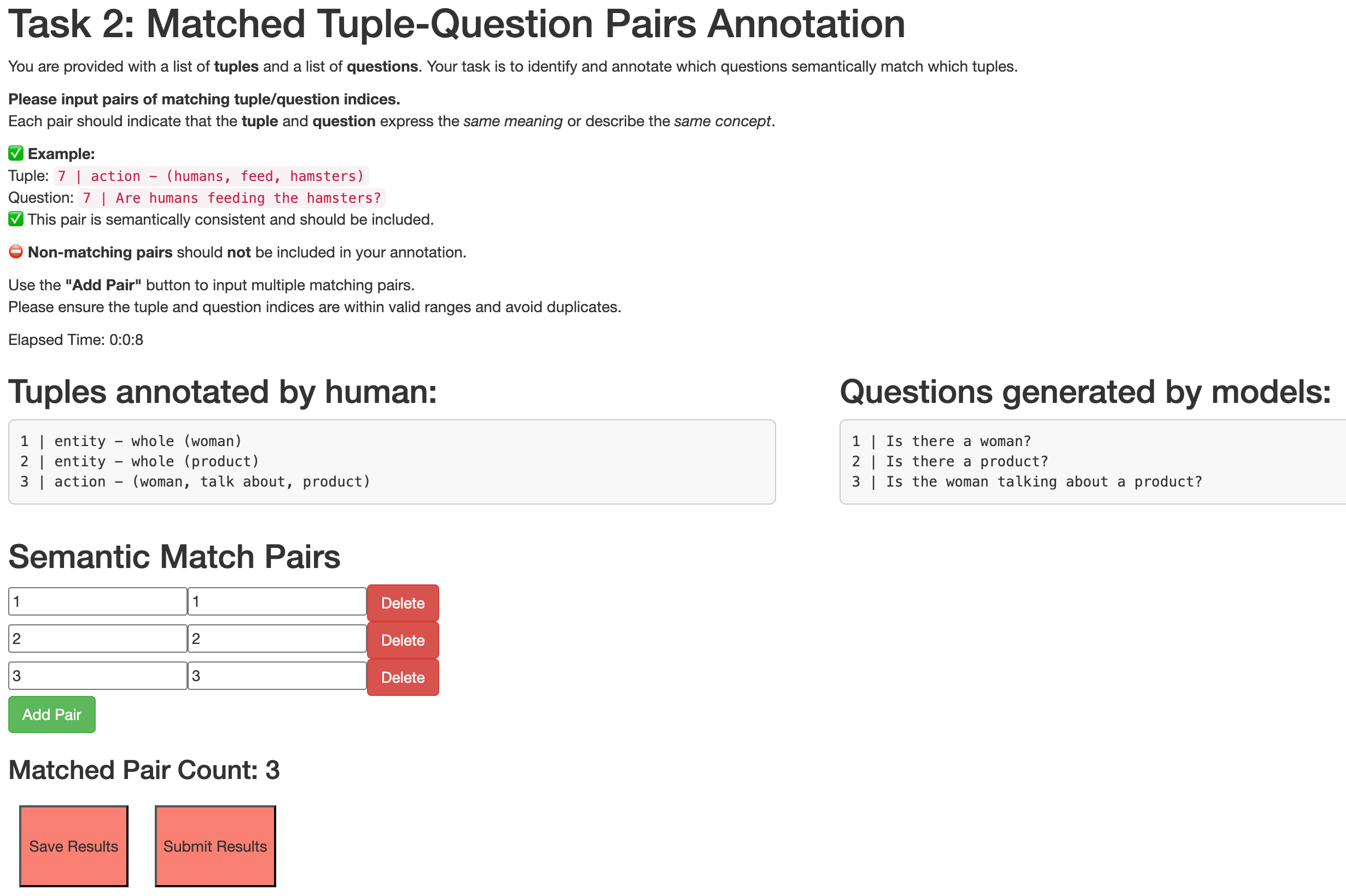}
    \caption{UI for question quality evaluation of human annotation.}
    \label{fig:t2_ui}
\end{figure*}

\begin{figure*}
    \centering
    \includegraphics[width=\linewidth]{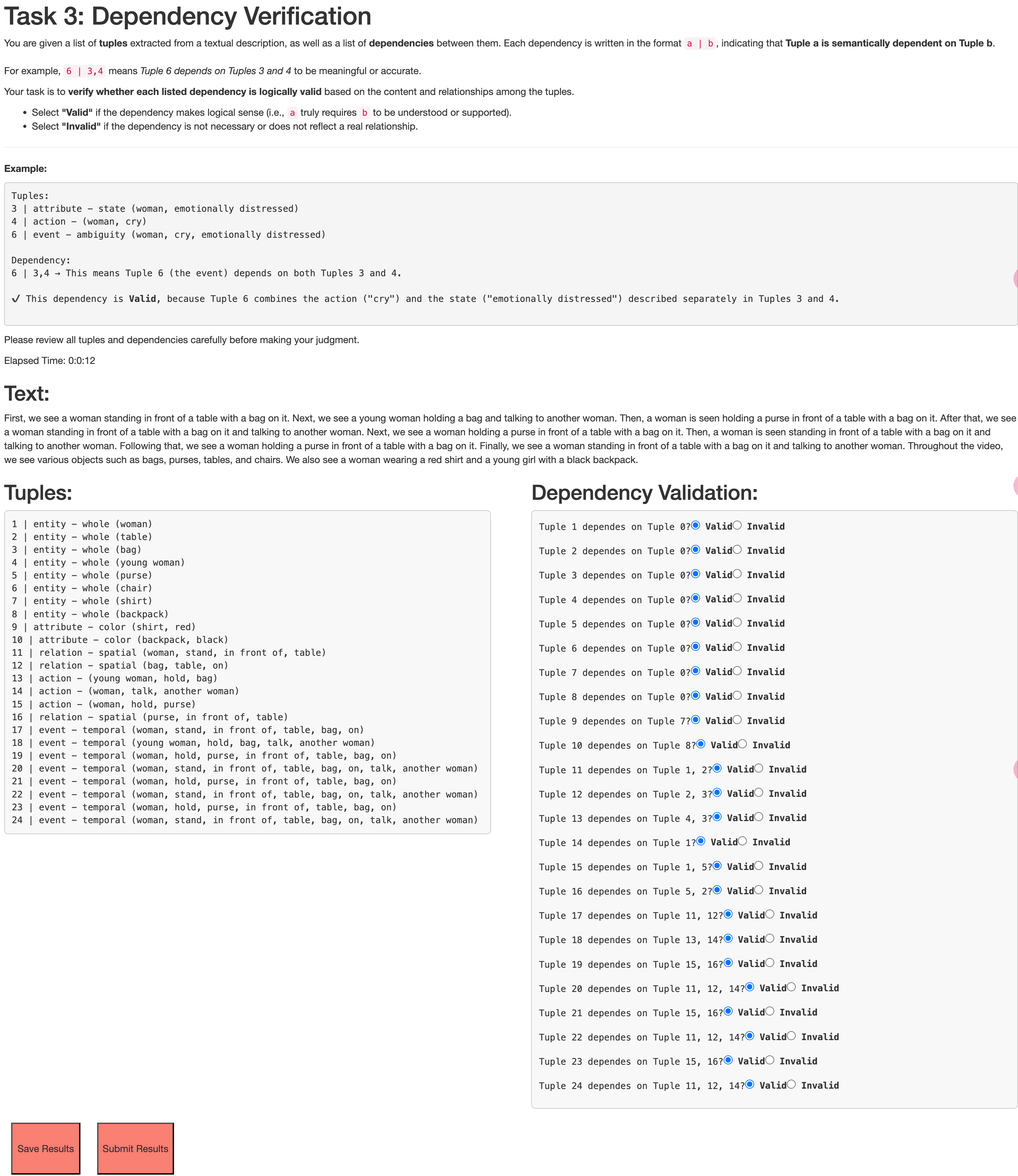}
    \caption{UI for dependency verification of human annotation.}
    \label{fig:t3_ui}
\end{figure*}

\begin{figure*}
    \centering
    \includegraphics[width=\linewidth]{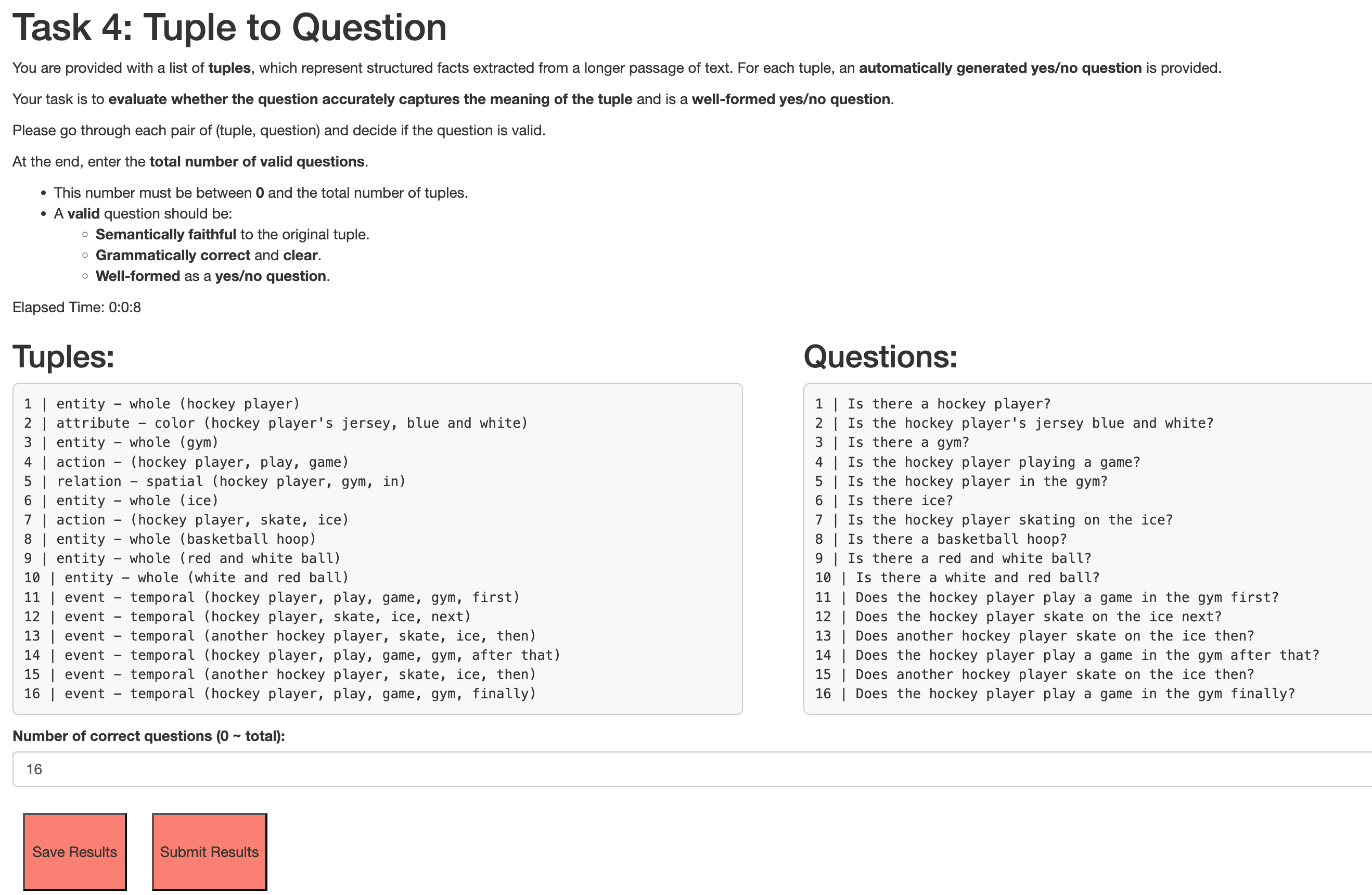}
    \caption{UI for the fact-to-question task of human evaluation.}
    \label{fig:t4_ui}
\end{figure*}

\begin{figure*}
    \centering
    \includegraphics[width=\linewidth]{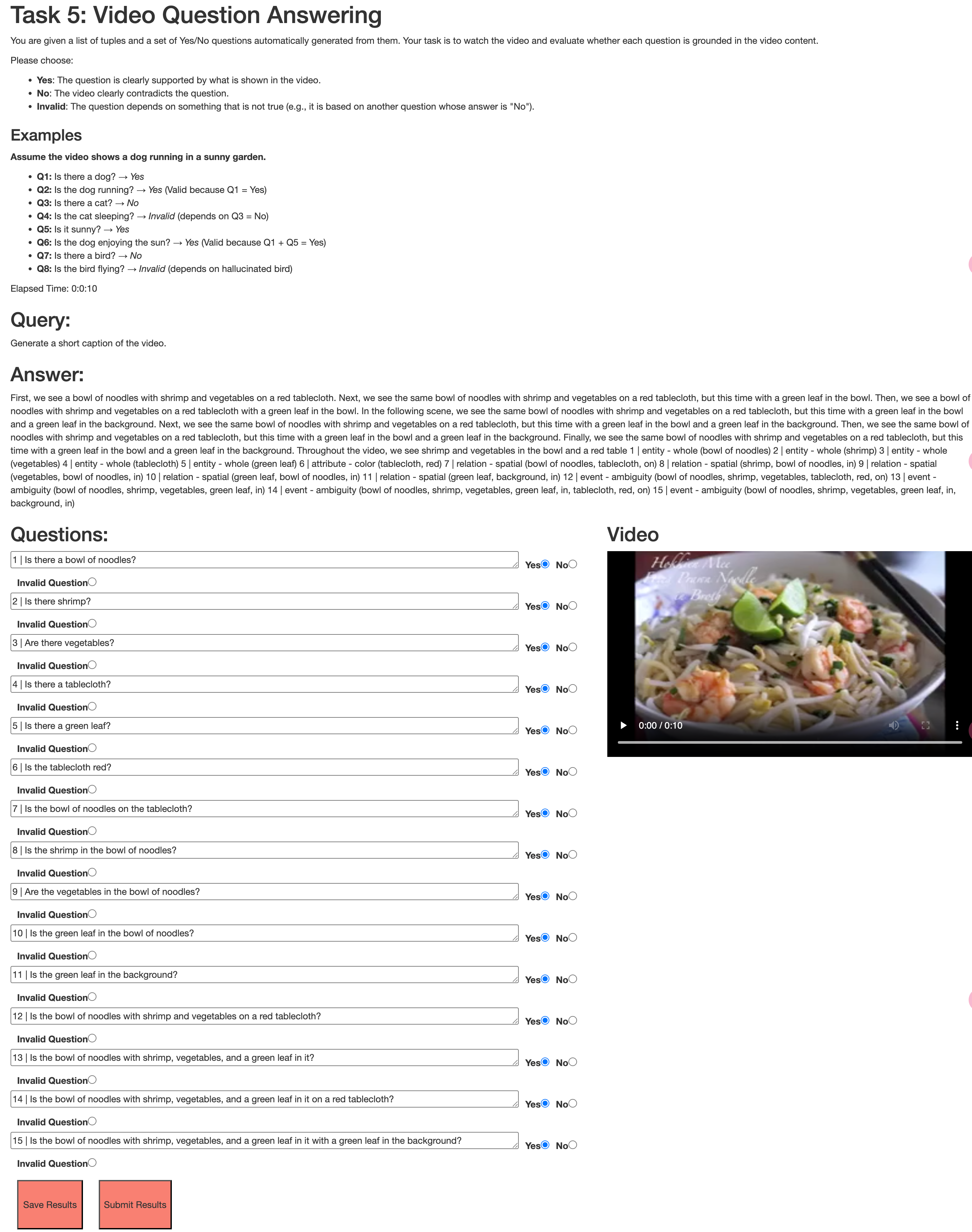}
    \caption{UI for Video Question Answering of human annotation.}
    \label{fig:t5_ui}
\end{figure*}

We show UI for all human evaluation tasks in Figure \ref{fig:t1_ui}, Figure \ref{fig:t2_ui}, Figure \ref{fig:t3_ui}, Figure \ref{fig:t4_ui}, and Figure \ref{fig:t5_ui}.

%% file: sections/tables/ICL-fact-extraction-T2V.tex
\section{In-context Examples for Prompts}
\label{appendix:ICL}
\subsection{Fact Extraction for Video-to-Text}
\begin{tiny}
\begin{lstlisting}
query: Please generate a caption for the video.
input: A male skateboarder is trying to pull off a trick on the ramp.
output: 1 | entity - whole (skateboarder)
2 | entity - whole (ramp)
3 | attribute - type (skateboarder, male)
4 | action - (male skateboarder, pull off a trick)
5 | relation - spatial (male skateboarder, ramp, on)
6 | event - ambigity (skateboarder, male, pull off a trick)
7 | event - ambiguity (male skateboarder, ramp, on)
8 | event - ambiguity (skateboarder, pull off a trick, ramp, on)

query: Please generate a caption for the video.
input: A car playing soccer, digital art.
output: 1 | entity - whole (car)
2 | global - (digital art)
3 | action - (car, soccer, play)

query: Please generate a caption for the video.
input: A set of 2x2 emoji icons with happy, angry, surprised and sobbing faces. The emoji icons look like pigs. All of the pigs are wearing crowns.
output: 1 | entity - whole (emoji icons)
2 | other - count (emoji icons, ==4)
3 | attribute - state (emoji icons, 2x2 grid)
4 | attribute - type (emoji icons, pig)
5 | attribute - state (emoji_1, happy)
6 | attribute - state (emoji_2, angry)
7 | attribute - state (emoji_3, surprised)
8 | attribute - state (emoji_4, sobbing face)
9 | entity - part (pig's crown)

query: Please generate a caption for the video.
input: a photo of bear and dining table; dining table is below bear
output: 1 | global - (photo)
2 | entity - whole (bear)
3 | entity - whole (dining table)
4 | relation - spatial (dining table, bear, below)

query: Please generate a caption for the video.
input: A group of children sitting in the grass with two of them holding a Frisbee .
output: 1 | entity - whole (children)
2 | entity - whole (grass)
3 | entity - whole (frisbee)
4 | attribute - state (children, sit)
5 | relation - spatial (a group of children, grass, sitting in)
6 | entity - part (two of the children)
7 | action - (two of the children, frisbee, hold)

query: Please generate a caption for the video.
input: the word 'START' written in chalk on a sidewalk
output: 1 | entity - whole (word)
2 | entity - whole (sidewalk)
3 | other - text rendering (word, "START")
4 | attribute - texture (word, chalk)
5 | relation - spatial (word 'START', sidewalk, on)

query: Please generate a caption for the video.
input: A pear, orange, and two bananas in a wooden bowl.
output: 1 | entity - whole (pear)
2 | entity - whole (orange)
3 | entity - whole (bananas)
4 | other - count (bananas, ==2)
5 | entity - whole (bowl)
6 | attribute - material (bowl, wood)
7 | relation - spatial (pear, bowl, in)
8 | relation - spatial (orange, bowl, in)
9 | relation - spatial (bananas, bowl, in)
10 | relation - spatial (bananas, bowl, in)
11 | event - ambiguity (pear, orange, bananas, ==2, bowl, in)

query: Please generate a caption for the video.
input: Closeup picture of the front of a clean motorcycle.
output: 1 | entity - whole (motorcycle)
2 | global - (closeup)
3 | global - (picture)
4 | attribute - state (motorcycle, clean)
5 | entity - part (front of the clean motorcycle)

query: Please generate a caption for the video.
input: a sad man with green hair
output: 1 | entity - whole (man)
2 | entity - part (man's hair)
3 | attribute - state (man, sad)
4 | attribute - color (man's hair, green)
5 | event - ambiguity (man, sad, man's hair, green)

query: Please generate a caption for the video.
input: A commercial airplane with propellers flying through the air.
output: 1 | entity - whole (airplane)
2 | entity - part (airplane's propellers)
3 | action - (airplane, air, fly through)
4 | event - ambiguity (airplane, with propellers, air, fly through)

query: Please generate a caption for the video.
input: A little boy grips a soccer ball in his arms surrounded by other youth soccer players.
output: 1 | entity - whole (boy)
2 | entity - whole (ball)
3 | entity - whole (soccer players)
4 | entity - part (boy's arms)
5 | entity - scale (boy, little)
6 | attribute - type (ball, soccer)
7 | attribute - state (soccer players, youth)
8 | relation - spatial (little boy, ball, grip in his arms)
9 | relation - spatial (little boy gripping the ball in his arms, soccer players, surrounded by)
10 | event - ambiguity (boy's arm, little, ball, soccer, grip in his arms)
11 | event - ambiguity (boy, little, soccer players, youth, surrounded by)

query: Please generate a caption for the video.
input: A traffic light and a signpost at a crossroads intersection near a waterway.
output: 1 | entity - whole (traffic light)
2 | entity - whole (signpost)
3 | entity - whole (crossroads intersection)
4 | entity - whole (waterway)
5 | relation - spatial (traffic light, crossroads intersection, at)
6 | relation - spatial (signpost, crossroads intersection, at)
7 | relation - spatial (traffic light, waterway, near)
8 | relation - spatial (signpost, waterway, near)
9 | relation - spatial (crossroads intersection, waterway, near)
10 | event - ambiguity (traffic light, signpost, crossroads intersection, at)
11 | event - ambiguity (traffic light, crossroads intersection, at, waterway, near)
12 | event - spatial (signpost, crossroads intersection, at, waterway, near)

query: Please generate a caption for the video.
input: a photo of dining table and traffic light; traffic light is below dining table
output: 1 | global - (photo)
2 | entity - whole (dining table)
3 | entity - whole (traffic light)
4 | relation - spatial (traffic light, dining table, below)

query: Please generate a caption for the video.
input: A realistic photo of a Pomeranian dressed up like a 1980s professional wrestler with neon green and neon orange face paint and bright green wrestling tights with bright orange boots.
output: 1 | global - (photo)
2 | entity - whole (Pomeranian)
3 | global - (realistic)
4 | entity - part (Pomeranian's costume)
5 | attribute - type (Pomeranian's costume, 1980s professional wrestler)
6 | entity - part (Pomeranian's costume's wrestling tights)
7 | entity - part (Pomeranian's costume's wrestling tights' boots)
8 | entity - part (Pomeranian's facepaint)
9 | attribute - color (Pomeranian's facepaint, neon green)
10 | attribute - color (Pomeranian's facepaint, neon orange)
11 | attribute - color (Pomeranian's costume's wrestling tights, bright green)
12 | attribute - color (Pomeranian's costume's wrestling tights' boots, bright orange)

query: Please generate a caption for the video.
input: a four-piece band on a stage in front of a small crowd
output: 1 | entity - whole (band)
2 | entity - whole (stage)
3 | entity - whole (crowd)
4 | other - count (band members, ==4)
5 | attribute - shape (crowd, small)
6 | relation - spatial (four-piece band, stage, on)
7 | relation - spatial (four-piece band, crowd, in front of)
8 | relation - spatial (stage, crowd, in front of)
9 | event - ambiguity (band, ==4 picece, stage, on)
10 | event - ambiguity (band, ==4 picece, crowd, small, in front of)
11 | event - ambiguity (stage, crowd, small, in front off)

query: Please generate a caption for the video.
input: two laptops, a mouse cord, and a monitor
output: 1 | entity - whole (laptops)
2 | other - count (laptops, ==2)
3 | entity - whole (mouse coord)
4 | entity - whole (monitor)

query: Please generate a caption for the video.
input: A red motorcycle parked by paint chipped doors.
output: 1 | entity - whole (motorcycle)
2 | entity - whole (doors)
3 | attribute - color (motorcycle, red)
4 | attribute - state (door, paint chipped)
5 | relation - spatial (red motorcycle, paint chipped door, next to)
6 | attribute - state (motorcycle, parked)
7 | event- ambiguity (motorcycle, red, door, paint chipped, next to)
8 | event- ambiguity (motorcycle, red, parked)

query: Please generate a caption for the video.
input: A cube made of denim. A cube with the texture of denim.
output: 1 | entity - whole (cube)
2 | attribute - material (cube, denim)
3 | attribute - texture (cube, denim)

query: Please generate a caption for the video.
input: an espresso machine that makes coffee from human souls
output: 1 | entity - whole (espresso machine)
2 | entity - whole (coffee)
3 | entity - whole (human souls)
4 | action - (espresso machine, coffee, make)
5 | attribute - material (coffee, human souls)
6 | event - ambiguity (espresso machine, coffee, make, human souls)

query: Please generate a caption for the video.
input: Three people standing next to an elephant along a river.
output: 1 | entity - whole (people)
2 | other - count (people, ==3)
3 | entity - whole (elephant)
4 | entity - whole (river)
5 | attribute - state (people, stand)
6 | relation - spatial (three people, elephant, next to)
7 | relation - spatial (people, river, next to)
8 | relation - spatial (elephant, river, next to)
9 | event - ambiguity (people, ==3, stand)
10 | event - ambiguity (people, ==3, elephant, next to)
11 | event - ambiguity (people, ==3, river, next to)
12 | event - ambiguity (people, stand, elephant, next to)
13 | event - ambiguity (people, stand, river, next to)
14 | event - ambiguity (people, elephant, next to, river, next to)

query: Please generate a caption for the video.
input: Aerial view of downtown Manhattan, but with Millennium Wheel next to the Statue of Liberty. The Great Pyramid is on a sandy island near the buildings.
output: 1 | entity - (downtown Manhattan)
2 | entity - (Millennium Wheel)
3 | entity - (the Statue of the Liberty)
4 | entity - (the Great Pyramid)
5 | entity - (island)
6 | entity - (buildings)
7 | global - (aerial view)
8 | attribute - texture (island, sandy)
9 | relation - spatial (Millennium Wheel, the Statue of Liberty, next to)
10 | relation - spatial (the Great Pyramid, island, on)
11 | relation - spatial (the Great Pyramid, buildings, near)
12 | event - ambiguity (the Great Pyramid, island, on, buildings, near)

query: Please generate a caption for the video.
input: A laptop with external keyboard, mouse, phone and photo on a desk.
output: 1 | entity - whole (laptop)
2 | entity - whole (keyboard)
3 | entity - whole (mouse)
4 | entity - whole (phone)
5 | entity - whole (photo)
6 | entity - whole (desk)
7 | attribute - type (keyboard, external)
8 | relation - spatial (laptop, desk, on)
9 | relation - spatial (keyboard, desk, on)
10 | relation - spatial (mouse, desk, on)
11 | relation - spatial (phone, desk, on)
12 | relation - spatial (photo, desk, on)
13 | event - ambiguity (laptop, external keyboard, mouse, phone, photo, desk, on)

query: Please generate a caption for the video.
input: A white slope covers the background, while the foreground features a grassy slope with several rams grazing and one measly and underdeveloped evergreen in the foreground.
output: 1 | entity - whole (slopes)
2 | other - count (slopes, ==2)
3 | entity - whole (rams)
4 | entity - whole (evergreen)
5 | attribute - color (slope_1, white)
6 | attribute - texture (slope_2, grassy)
7 | attribute - state (evergreen, measly and underdeveloped)
8 | relation - spatial (slope_1, background, in)
9 | relation - spatial (slope_2, foreground, in)
10 | relation - spatial (several, rams, grassy slope_2, on)
11 | attribute - state (several rams, graze)
12 | event - ambiguity (slope_1, white, background, in)
13 | event - ambiguity (slope_2, grassy, foreground, in)
14 | event - ambiguity (several, rams, slope_2, grassy, on)

query: Please generate a caption for the video.
input: A man walks into a room and sits on a chair. A dog follows him.
output: 1 | entity - whole (man)
2 | entity - whole (room)
3 | entity - whole (chair)
4 | entity - whole (dog)
5 | action - (man, walk, room)
6 | action - (man, sit on, chair)
7 | action - (dog, follow, man)
8 | relation - temporal (man, sit, before, walk)
9 | relation - temporal (dog, follows, after, man, sit)
10 | event - temporal (man, walks into a room and sits on a chair, dog follows him)

query: Please generate a caption for the video.
input: A car is parked by the roadside. Later, it starts moving and drives away.
output: 1 | entity - whole (car)
2 | entity - whole (roadside)
3 | relation - spatial (car, roadside, park)
4 | action - (car, move)
5 | action - (car, drives away)
6 | relation - temporal (car, starts, after, parked)
7 | relation - temporal (car, drive away, after, parked)
8 | event - temporal (car, move, roadside, park, after)
9 | event - temporal (car, drive away, roadside, park, after)
10 | event - temporal (car, starts, parked, move, drive away)

query: What's unusual in this video?
input: A man is running across a street while carrying a large bag. This is unusual because people typically do not carry large bags while running across streets.
output: 1 | entity - whole (man)
2 | entity - whole (street)
3 | entity - whole (bag)
4 | relation - spatial (man, run, street)
5 | entity - scale (large bag)
6 | relation - spatial (man, carry, large bag)
7 | relation - temporal (man, carry, while, running)
8 | event - ambiguity (man, large bag, carry)
9 | event - temporal (man, run, street, while, carry, large bag)
\end{lstlisting}
\end{tiny}

\subsection{Fact Extraction for Text-to-Video}
\begin{tiny}
\begin{lstlisting}
input: A male skateboarder is trying to pull off a trick on the ramp.
output: 1 | entity - whole (skateboarder)
2 | entity - whole (ramp)
3 | attribute - type (skateboarder, male)
4 | action - (male skateboarder, pull off a trick)
5 | relation - spatial (male skateboarder, ramp, on)
6 | event - ambigity (skateboarder, male, pull off a trick)
7 | event - ambiguity (male skateboarder, ramp, on)
8 | event - ambiguity (skateboarder, pull off a trick, ramp, on)

input: A car playing soccer, digital art.
output: 1 | entity - whole (car)
2 | global - (digital art)
3 | action - (car, soccer, play)

input: A set of 2x2 emoji icons with happy, angry, surprised and sobbing faces. The emoji icons look like pigs. All of the pigs are wearing crowns.
output: 1 | entity - whole (emoji icons)
2 | other - count (emoji icons, ==4)
3 | attribute - state (emoji icons, 2x2 grid)
4 | attribute - type (emoji icons, pig)
5 | attribute - state (emoji_1, happy)
6 | attribute - state (emoji_2, angry)
7 | attribute - state (emoji_3, surprised)
8 | attribute - state (emoji_4, sobbing face)
9 | entity - part (pig's crown)

input: a photo of bear and dining table; dining table is below bear
output: 1 | global - (photo)
2 | entity - whole (bear)
3 | entity - whole (dining table)
4 | relation - spatial (dining table, bear, below)

input: A group of children sitting in the grass with two of them holding a Frisbee .
output: 1 | entity - whole (children)
2 | entity - whole (grass)
3 | entity - whole (frisbee)
4 | attribute - state (children, sit)
5 | relation - spatial (a group of children, grass, sitting in)
6 | entity - part (two of the children)
7 | action - (two of the children, frisbee, hold)

input: the word 'START' written in chalk on a sidewalk
output: 1 | entity - whole (word)
2 | entity - whole (sidewalk)
3 | other - text rendering (word, "START")
4 | attribute - texture (word, chalk)
5 | relation - spatial (word 'START', sidewalk, on)

input: A pear, orange, and two bananas in a wooden bowl.
output: 1 | entity - whole (pear)
2 | entity - whole (orange)
3 | entity - whole (bananas)
4 | other - count (bananas, ==2)
5 | entity - whole (bowl)
6 | attribute - material (bowl, wood)
7 | relation - spatial (pear, bowl, in)
8 | relation - spatial (orange, bowl, in)
9 | relation - spatial (bananas, bowl, in)
10 | relation - spatial (bananas, bowl, in)
11 | event - ambiguity (pear, orange, bananas, ==2, bowl, in)

input: Closeup picture of the front of a clean motorcycle.
output: 1 | entity - whole (motorcycle)
2 | global - (closeup)
3 | global - (picture)
4 | attribute - state (motorcycle, clean)
5 | entity - part (front of the clean motorcycle)

input: a sad man with green hair
output: 1 | entity - whole (man)
2 | entity - part (man's hair)
3 | attribute - state (man, sad)
4 | attribute - color (man's hair, green)
5 | event - ambiguity (man, sad, man's hair, green)

input: A commercial airplane with propellers flying through the air.
output: 1 | entity - whole (airplane)
2 | entity - part (airplane's propellers)
3 | action - (airplane, air, fly through)
4 | event - ambiguity (airplane, with propellers, air, fly through)

input: A little boy grips a soccer ball in his arms surrounded by other youth soccer players.
output: 1 | entity - whole (boy)
2 | entity - whole (ball)
3 | entity - whole (soccer players)
4 | entity - part (boy's arms)
5 | entity - scale (boy, little)
6 | attribute - type (ball, soccer)
7 | attribute - state (soccer players, youth)
8 | relation - spatial (little boy, ball, grip in his arms)
9 | relation - spatial (little boy gripping the ball in his arms, soccer players, surrounded by)
10 | event - ambiguity (boy's arm, little, ball, soccer, grip in his arms)
11 | event - ambiguity (boy, little, soccer players, youth, surrounded by)

input: A traffic light and a signpost at a crossroads intersection near a waterway.
output: 1 | entity - whole (traffic light)
2 | entity - whole (signpost)
3 | entity - whole (crossroads intersection)
4 | entity - whole (waterway)
5 | relation - spatial (traffic light, crossroads intersection, at)
6 | relation - spatial (signpost, crossroads intersection, at)
7 | relation - spatial (traffic light, waterway, near)
8 | relation - spatial (signpost, waterway, near)
9 | relation - spatial (crossroads intersection, waterway, near)
10 | event - ambiguity (traffic light, signpost, crossroads intersection, at)
11 | event - ambiguity (traffic light, crossroads intersection, at, waterway, near)
12 | event - spatial (signpost, crossroads intersection, at, waterway, near)

input: a photo of dining table and traffic light; traffic light is below dining table
output: 1 | global - (photo)
2 | entity - whole (dining table)
3 | entity - whole (traffic light)
4 | relation - spatial (traffic light, dining table, below)

input: A realistic photo of a Pomeranian dressed up like a 1980s professional wrestler with neon green and neon orange face paint and bright green wrestling tights with bright orange boots.
output: 1 | global - (photo)
2 | entity - whole (Pomeranian)
3 | global - (realistic)
4 | entity - part (Pomeranian's costume)
5 | attribute - type (Pomeranian's costume, 1980s professional wrestler)
6 | entity - part (Pomeranian's costume's wrestling tights)
7 | entity - part (Pomeranian's costume's wrestling tights' boots)
8 | entity - part (Pomeranian's facepaint)
9 | attribute - color (Pomeranian's facepaint, neon green)
10 | attribute - color (Pomeranian's facepaint, neon orange)
11 | attribute - color (Pomeranian's costume's wrestling tights, bright green)
12 | attribute - color (Pomeranian's costume's wrestling tights' boots, bright orange)

input: a four-piece band on a stage in front of a small crowd
output: 1 | entity - whole (band)
2 | entity - whole (stage)
3 | entity - whole (crowd)
4 | other - count (band members, ==4)
5 | attribute - shape (crowd, small)
6 | relation - spatial (four-piece band, stage, on)
7 | relation - spatial (four-piece band, crowd, in front of)
8 | relation - spatial (stage, crowd, in front of)
9 | event - ambiguity (band, ==4 picece, stage, on)
10 | event - ambiguity (band, ==4 picece, crowd, small, in front of)
11 | event - ambiguity (stage, crowd, small, in front off)

input: two laptops, a mouse cord, and a monitor
output: 1 | entity - whole (laptops)
2 | other - count (laptops, ==2)
3 | entity - whole (mouse coord)
4 | entity - whole (monitor)

input: A red motorcycle parked by paint chipped doors.
output: 1 | entity - whole (motorcycle)
2 | entity - whole (doors)
3 | attribute - color (motorcycle, red)
4 | attribute - state (door, paint chipped)
5 | relation - spatial (red motorcycle, paint chipped door, next to)
6 | attribute - state (motorcycle, parked)
7 | event- ambiguity (motorcycle, red, door, paint chipped, next to)
8 | event- ambiguity (motorcycle, red, parked)

input: A cube made of denim. A cube with the texture of denim.
output: 1 | entity - whole (cube)
2 | attribute - material (cube, denim)
3 | attribute - texture (cube, denim)

input: an espresso machine that makes coffee from human souls
output: 1 | entity - whole (espresso machine)
2 | entity - whole (coffee)
3 | entity - whole (human souls)
4 | action - (espresso machine, coffee, make)
5 | attribute - material (coffee, human souls)
6 | event - ambiguity (espresso machine, coffee, make, human souls)

input: Three people standing next to an elephant along a river.
output: 1 | entity - whole (people)
2 | other - count (people, ==3)
3 | entity - whole (elephant)
4 | entity - whole (river)
5 | attribute - state (people, stand)
6 | relation - spatial (three people, elephant, next to)
7 | relation - spatial (people, river, next to)
8 | relation - spatial (elephant, river, next to)
9 | event - ambiguity (people, ==3, stand)
10 | event - ambiguity (people, ==3, elephant, next to)
11 | event - ambiguity (people, ==3, river, next to)
12 | event - ambiguity (people, stand, elephant, next to)
13 | event - ambiguity (people, stand, river, next to)
14 | event - ambiguity (people, elephant, next to, river, next to)

input: Aerial view of downtown Manhattan, but with Millennium Wheel next to the Statue of Liberty. The Great Pyramid is on a sandy island near the buildings.
output: 1 | entity - (downtown Manhattan)
2 | entity - (Millennium Wheel)
3 | entity - (the Statue of the Liberty)
4 | entity - (the Great Pyramid)
5 | entity - (island)
6 | entity - (buildings)
7 | global - (aerial view)
8 | attribute - texture (island, sandy)
9 | relation - spatial (Millennium Wheel, the Statue of Liberty, next to)
10 | relation - spatial (the Great Pyramid, island, on)
11 | relation - spatial (the Great Pyramid, buildings, near)
12 | event - ambiguity (the Great Pyramid, island, on, buildings, near)

input: A laptop with external keyboard, mouse, phone and photo on a desk.
output: 1 | entity - whole (laptop)
2 | entity - whole (keyboard)
3 | entity - whole (mouse)
4 | entity - whole (phone)
5 | entity - whole (photo)
6 | entity - whole (desk)
7 | attribute - type (keyboard, external)
8 | relation - spatial (laptop, desk, on)
9 | relation - spatial (keyboard, desk, on)
10 | relation - spatial (mouse, desk, on)
11 | relation - spatial (phone, desk, on)
12 | relation - spatial (photo, desk, on)
13 | event - ambiguity (laptop, external keyboard, mouse, phone, photo, desk, on)

input: A white slope covers the background, while the foreground features a grassy slope with several rams grazing and one measly and underdeveloped evergreen in the foreground.
output: 1 | entity - whole (slopes)
2 | other - count (slopes, ==2)
3 | entity - whole (rams)
4 | entity - whole (evergreen)
5 | attribute - color (slope_1, white)
6 | attribute - texture (slope_2, grassy)
7 | attribute - state (evergreen, measly and underdeveloped)
8 | relation - spatial (slope_1, background, in)
9 | relation - spatial (slope_2, foreground, in)
10 | relation - spatial (several, rams, grassy slope_2, on)
11 | attribute - state (several rams, graze)
12 | event - ambiguity (slope_1, white, background, in)
13 | event - ambiguity (slope_2, grassy, foreground, in)
14 | event - ambiguity (several, rams, slope_2, grassy, on)

input: A man walks into a room and sits on a chair. A dog follows him.
output: 1 | entity - whole (man)
2 | entity - whole (room)
3 | entity - whole (chair)
4 | entity - whole (dog)
5 | action - (man, walk, room)
6 | action - (man, sit on, chair)
7 | action - (dog, follow, man)
8 | relation - temporal (man, sit, before, walk)
9 | relation - temporal (dog, follows, after, man, sit)
10 | event - temporal (man, walks into a room and sits on a chair, dog follows him)

input: A car is parked by the roadside. Later, it starts moving and drives away.
output: 1 | entity - whole (car)
2 | entity - whole (roadside)
3 | relation - spatial (car, roadside, park)
4 | action - (car, move)
5 | action - (car, drives away)
6 | relation - temporal (car, starts, after, parked)
7 | relation - temporal (car, drive away, after, parked)
8 | event - temporal (car, move, roadside, park, after)
9 | event - temporal (car, drive away, roadside, park, after)
10 | event - temporal (car, starts, parked, move, drive away)

input: A man is running across a street while carrying a large bag. This is unusual because people typically do not carry large bags while running across streets.
output: 1 | entity - whole (man)
2 | entity - whole (street)
3 | entity - whole (bag)
4 | relation - spatial (man, run, street)
5 | entity - scale (large bag)
6 | relation - spatial (man, carry, large bag)
7 | relation - temporal (man, carry, while, running)
8 | event - ambiguity (man, large bag, carry)
9 | event - temporal (man, run, street, while, carry, large bag)
\end{lstlisting}
\end{tiny}

\subsection{Question Generation}
\begin{tiny}
\begin{lstlisting}
input: A male skateboarder is trying to pull off a trick on the ramp. 
1 | entity - whole (skateboarder)
2 | entity - whole (ramp)
3 | attribute - type (skateboarder, male)
4 | action - (male skateboarder, pull off a trick)
5 | relation - spatial (male skateboarder, ramp, on)
6 | event - ambigity (skateboarder, male, pull off a trick)
7 | event - ambiguity (male skateboarder, ramp, on)
8 | event - ambiguity (skateboarder, pull off a trick, ramp, on)
output: 1 | Is there a skateboarder?
2 | Is there a ramp?
3 | Is the skateboarder male?
4 | Is the skateboarder pulling off a trick?
5 | Is the skateboarder on the ramp?
6 | Is the male skateboarder on the ramp?
7 | Is the male skateboarder on the ramp?
8 | Is the skateboarder pulling off a trick on the ramp?

input: A car playing soccer, digital art.
1 | entity - whole (car)
2 | global - (digital art)
3 | action - (car, soccer, play)
output: 1 | Is there a car?
2 | Is this digital art?
3 | Is the car playing soccer?

input: A set of 2x2 emoji icons with happy, angry, surprised and sobbing faces. The emoji icons look like pigs. All of the pigs are wearing crowns.
1 | entity - whole (emoji icons)
2 | other - count (emoji icons, ==4)
3 | attribute - state (emoji icons, 2x2 grid)
4 | attribute - type (emoji icons, pig)
5 | attribute - state (emoji_1, happy)
6 | attribute - state (emoji_2, angry)
7 | attribute - state (emoji_3, surprised)
8 | attribute - state (emoji_4, sobbing face)
9 | entity - part (pig's crown)
output: 1 | nan
2 | Is there a total of four emoji icons?
3 | Were the emojis in a 2x2 grid?
4 | Did emojis look like pigs?
5 | Did one emoji look happy?
6 | Did one emoji look angry?
7 | Did one emoji look surprised?
8 | Did the emoji have a sobbing face?
9 | Are all the emoji wearing crowns?

input: a photo of bear and dining table; dining table is below bear
1 | global - (photo)
2 | entity - whole (bear)
3 | entity - whole (dining table)
4 | relation - spatial (dining table, bear, below)
output: 1 | Is this a photo?
2 | Is there a bear?
3 | Is there a dining table?
4 | Is the dining table below the bear?

input: A group of children sitting in the grass with two of them holding a Frisbee .
1 | entity - whole (children)
2 | entity - whole (grass)
3 | entity - whole (frisbee)
4 | attribute - state (children, sit)
5 | relation - spatial (a group of children, grass, sitting in)
6 | entity - part (two of the children)
7 | action - (two of the children, frisbee, hold)
output: 1 | Are there a group of children?
2 | Is there grass?
3 | Is there a frisbee?
4 | Are the children sitting?
5 | Are a group of children sitting in the grass?
6 | Are there two of the children?
7 | Are two of the children holding a frisbee?

input: the word 'START' written in chalk on a sidewalk
1 | entity - whole (word)
2 | entity - whole (sidewalk)
3 | other - text rendering (word, "START")
4 | attribute - texture (word, chalk)
5 | relation - spatial (word 'START', sidewalk, on)
output: 1 | Is there a word?
2 | Is there a sidewalk?
3 | Does the word say "START"?
4 | Is the word written in chalk?
5 | Is the word 'START' on the sidewalk?

input: A pear, orange, and two bananas in a wooden bowl.
1 | entity - whole (pear)
2 | entity - whole (orange)
3 | entity - whole (bananas)
4 | other - count (bananas, ==2)
5 | entity - whole (bowl)
6 | attribute - material (bowl, wood)
7 | relation - spatial (pear, bowl, in)
8 | relation - spatial (orange, bowl, in)
9 | relation - spatial (bananas, bowl, in)
10 | relation - spatial (bananas, bowl, in)
11 | event - ambiguity (pear, orange, bananas, ==2, bowl, in)
output: 1 | Is there a pear?
2 | Is there an orange?
3 | Are there bananas?
4 | Are there two bananas?
5 | Is there a bowl?
6 | Is the bowl made of wood?
7 | Is the pear in the wooden bowl?
8 | Is the orange in the wooden bowl?
9 | Are bananas in the wooden bowl?
10 | Are bananas in the wooden bowl?
11 | Are the pear, the orange and two bananas bananas in the same wooden bowl?

input: Closeup picture of the front of a clean motorcycle.
1 | entity - whole (motorcycle)
2 | global - (closeup)
3 | global - (picture)
4 | attribute - state (motorcycle, clean)
5 | entity - part (front of the clean motorcycle)
output: 1 | Is there a motorcycle?
2 | Is this a closeup image?
3 | Is this a picture?
4 | Is the motorcycle clean?
5 | Is the closeup picture in the front of the clean motorcycle?

input: a sad man with green hair
1 | entity - whole (man)
2 | entity - part (man's hair)
3 | attribute - state (man, sad)
4 | attribute - color (man's hair, green)
5 | event - ambiguity (man, sad, man's hair, green)
output: 1 | Is there a man?
2 | Is there hair?
3 | Was the man sad?
4 | Is the hair green?
5 | Is the sad man with hair green?

input: A commercial airplane with propellers flying through the air.
1 | entity - whole (airplane)
2 | entity - part (airplane's propellers)
3 | action - (airplane, air, fly through)
4 | event - ambiguity (airplane, with propellers, air, fly through)
output: 1 | Is there an airplane?
2 | Does the airplane have propellers?
3 | Is the airplane flying through the air?
4 | Is the airplane with propellers flying through the air?

input: A little boy grips a soccer ball in his arms surrounded by other youth soccer players.
1 | entity - whole (boy)
2 | entity - whole (ball)
3 | entity - whole (soccer players)
4 | entity - part (boy's arms)
5 | entity - scale (boy, little)
6 | attribute - type (ball, soccer)
7 | attribute - state (soccer players, youth)
8 | relation - spatial (little boy, ball, grip in his arms)
9 | relation - spatial (little boy gripping the ball in his arms, soccer players, surrounded by)
10 | event - ambiguity (boy's arm, little, ball, soccer, grip in his arms)
11 | event - ambiguity (boy, little, soccer players, youth, surrounded by)
output: 1 | Is there a boy?
2 | Is there a ball?
3 | Are there other soccer players?
4 | Does the boy have arms?
5 | Is the boy little?
6 | Is the ball a soccer ball?
7 | Are the other soccer players young?
8 | Is the boy gripping the ball in his arms?
9 | Is the little boy surrounded by the other soccer players?
10 | Is the little boy gripping the soccer ball in his arms?
11 | Is the little boy surrounded by the other youth soccer players?

input: A traffic light and a signpost at a crossroads intersection near a waterway.
1 | entity - whole (traffic light)
2 | entity - whole (signpost)
3 | entity - whole (crossroads intersection)
4 | entity - whole (waterway)
5 | relation - spatial (traffic light, crossroads intersection, at)
6 | relation - spatial (signpost, crossroads intersection, at)
7 | relation - spatial (traffic light, waterway, near)
8 | relation - spatial (signpost, waterway, near)
9 | relation - spatial (crossroads intersection, waterway, near)
10 | event - ambiguity (traffic light, signpost, crossroads intersection, at)
11 | event - ambiguity (traffic light, crossroads intersection, at, waterway, near)
12 | event - spatial (signpost, crossroads intersection, at, waterway, near)
output: 1 | Is there a light?
2 | Is there a signpost?
3 | Is there an intersection?
4 | Is there a waterway?
5 | Is the light a traffic light?
6 | Is the intersection a crossroads intersection?
7 | Is the traffic light at the crossroads intersection?
8 | Is the signpost at the crossroads intersection?
9 | Is the intersection near the waterway?
10 | Are the traffic light and signpost at a crossrodas intersection?
11 | Is the traffic light at a crossrodas intersection near waterway?
12 | Is the signpost at a crossrodas intersection near waterway?

input: a photo of dining table and traffic light; traffic light is below dining table
1 | global - (photo)
2 | entity - whole (dining table)
3 | entity - whole (traffic light)
4 | relation - spatial (traffic light, dining table, below)
output: 1 | Is this a photo?
2 | Is there a dining table?
3 | Is there a traffic light?
4 | Is the traffice light below the dining table?

input: A realistic photo of a Pomeranian dressed up like a 1980s professional wrestler with neon green and neon orange face paint and bright green wrestling tights with bright orange boots.
1 | global - (photo)
2 | entity - whole (Pomeranian)
3 | global - (realistic)
4 | entity - part (Pomeranian's costume)
5 | attribute - type (Pomeranian's costume, 1980s professional wrestler)
6 | entity - part (Pomeranian's costume's wrestling tights)
7 | entity - part (Pomeranian's costume's wrestling tights' boots)
8 | entity - part (Pomeranian's facepaint)
9 | attribute - color (Pomeranian's facepaint, neon green)
10 | attribute - color (Pomeranian's facepaint, neon orange)
11 | attribute - color (Pomeranian's costume's wrestling tights, bright green)
12 | attribute - color (Pomeranian's costume's wrestling tights' boots, bright orange)
output: 1 | Is this a photo?
2 | Is there a Pomeranian?
3 | Is the photo realistic?
4 | Is the Pomeranian dressed up?
5 | Is the costume of a 1980s professional wrestler?
6 | Are wrestling tights included in the costume?
7 | Did the costume come with boots?
8 | Does the Pomeranian has a facepaint?
9 | Is the facepaint neon green?
10 | Is the facepaint neon orange?
11 | Are the wrestling tights bright green?
12 | Are the boots bright orange?

input: a four-piece band on a stage in front of a small crowd
1 | entity - whole (band)
2 | entity - whole (stage)
3 | entity - whole (crowd)
4 | other - count (band members, ==4)
5 | attribute - shape (crowd, small)
6 | relation - spatial (four-piece band, stage, on)
7 | relation - spatial (four-piece band, crowd, in front of)
8 | relation - spatial (stage, crowd, in front of)
9 | event - ambiguity (band, ==4 picece, stage, on)
10 | event - ambiguity (band, ==4 picece, crowd, small, in front of)
11 | event - ambiguity (stage, crowd, small, in front off)
output: 1 | Is there a band?
2 | Is there a stage?
3 | Is there a crowd?
4 | Is the band a fourpiece band?
5 | Is the crowd small?
6 | Is the band on the stage?
7 | Is the band in front of the crowd?
8 | Is the stage in front of the crowd?
9 | Are the four-piece band on the stage?
10 | Is the four-piece band in front of the small crowd?
11 | Is the stage in front of the small crowd?

input: two laptops, a mouse cord, and a monitor 
1 | entity - whole (laptops)
2 | other - count (laptops, ==2)
3 | entity - whole (mouse coord)
4 | entity - whole (monitor)
output: 1 | Are there laptops?
2 | Are there two laptops?
3 | Is there a cord?
4 | Is there a monitor?

input: A red motorcycle parked by paint chipped doors.
1 | entity - whole (motorcycle)
2 | entity - whole (doors)
3 | attribute - color (motorcycle, red)
4 | attribute - state (door, paint chipped)
5 | relation - spatial (red motorcycle, paint chipped door, next to)
6 | attribute - state (motorcycle, parked)
7 | event- ambiguity (motorcycle, red, door, paint chipped, next to)
8 | event- ambiguity (motorcycle, red, parked)
output: 1 | Is there a motorcycle?
2 | Are there any doors?
3 | Are the doors painted?
4 | Is the paint chipped?
5 | Is the motorcycle next to doors?
6 | Is the motorcycle parked?
7 | Is the red motorcycle next to paint chipped doors?
8 | Is the red motorcycle parked?

input: A cube made of denim. A cube with the texture of denim.
1 | entity - whole (cube)
2 | attribute - material (cube, denim)
3 | attribute - texture (cube, denim)
output: 1 | Is there a cube?
2 | Is the cube made of denim?
3 | Does the cube have texture of denim?

input: an espresso machine that makes coffee from human souls
1 | entity - whole (espresso machine)
2 | entity - whole (coffee)
3 | entity - whole (human souls)
4 | action - (espresso machine, coffee, make)
5 | attribute - material (coffee, human souls)
6 | event - ambiguity (espresso machine, coffee, make, human souls)
output: 1 | Do we have an espresso machine?
2 | Do we have coffee?
3 | Do human beings have souls?
4 | Is the espresso machine making coffee?
5 | Is the expersso made of human souls?
6 | Is the expersso machine making coffe with human souls?

input: Three people standing next to an elephant along a river.
1 | entity - whole (people)
2 | other - count (people, ==3)
3 | entity - whole (elephant)
4 | entity - whole (river)
5 | attribute - state (people, stand)
6 | relation - spatial (three people, elephant, next to)
7 | relation - spatial (people, river, next to)
8 | relation - spatial (elephant, river, next to)
9 | event - ambiguity (people, ==3, stand)
10 | event - ambiguity (people, ==3, elephant, next to)
11 | event - ambiguity (people, ==3, river, next to)
12 | event - ambiguity (people, stand, elephant, next to)
13 | event - ambiguity (people, stand, river, next to)
14 | event - ambiguity (people, elephant, next to, river, next to)
output: 1 | Are there people?
2 | Are there three people?
3 | Is there an elephant?
4 | Is there a river?
5 | Are people standing?
6 | Are people next to the elephant?
7 | Are people next to the river?
8 | Is the elephant next to the river?
9 | Are the three people standing?
10 | Are the three people next to the elephant?
11 | Are the three people next to the river?
12 | Are people stadning next to an elephant?
13 | Are people standing next to the river?
14 | Are people next to the river and an elephant?

input: Aerial view of downtown Manhattan, but with Millennium Wheel next to the Statue of Liberty. The Great Pyramid is on a sandy island near the buildings.
1 | entity - (downtown Manhattan)
2 | entity - (Millennium Wheel)
3 | entity - (the Statue of the Liberty)
4 | entity - (the Great Pyramid)
5 | entity - (island)
6 | entity - (buildings)
7 | global - (aerial view)
8 | attribute - texture (island, sandy)
9 | relation - spatial (Millennium Wheel, the Statue of Liberty, next to)
10 | relation - spatial (the Great Pyramid, island, on)
11 | relation - spatial (the Great Pyramid, buildings, near)
12 | event - ambiguity (the Great Pyramid, island, on, buildings, near)
output: 1 | Is downtown Manhattan there?
2 | Is Millennium Wheel there?
3 | Is the Statue of Liberty there?
4 | Is the Great Pyramid there?
5 | Is there an island?
6 | Are there buildings?
7 | Is this an aerial view?
8 | Is there the island sandy?
9 | Is the Millennium Wheel next to the Statue of Liberty?
10 | Is the Great Pyramid on the sandy island?
11 | Is the Great Pyramid near the buildings?
12 | Is the Great Pyramid on a sady island near the buildings?

input: A laptop with external keyboard, mouse, phone and photo on a desk.
1 | entity - whole (laptop)
2 | entity - whole (keyboard)
3 | entity - whole (mouse)
4 | entity - whole (phone)
5 | entity - whole (photo)
6 | entity - whole (desk)
7 | attribute - type (keyboard, external)
8 | relation - spatial (laptop, desk, on)
9 | relation - spatial (keyboard, desk, on)
10 | relation - spatial (mouse, desk, on)
11 | relation - spatial (phone, desk, on)
12 | relation - spatial (photo, desk, on)
13 | event - ambiguity (laptop, external keyboard, mouse, phone, photo, desk, on)
output: 1 | Is there a laptop?
2 | Is there a keyboard?
3 | Is there a mouse?
4 | Is there a phone?
5 | Is there a photo?
6 | Is there a desk?
7 | Is the keyboard external?
8 | Is the laptop on the desk?
9 | Is the keyboard on the desk?
10 | Is the mouse on the desk?
11 | Is the phone on the desk?
12 | Is the photo on the desk?
13 | Is all laptop, external keyboard, mouse, phone, photo on the same desk?

input: A white slope covers the background, while the foreground features a grassy slope with several rams grazing and one measly and underdeveloped evergreen in the foreground.  
1 | entity - whole (slopes)
2 | other - count (slopes, ==2)
3 | entity - whole (rams)
4 | entity - whole (evergreen)
5 | attribute - color (slope_1, white)
6 | attribute - texture (slope_2, grassy)
7 | attribute - state (evergreen, measly and underdeveloped)
8 | relation - spatial (slope_1, background, in)
9 | relation - spatial (slope_2, foreground, in)
10 | relation - spatial (several, rams, grassy slope_2, on)
11 | attribute - state (several rams, graze)
12 | event - ambiguity (slope_1, white, background, in)
13 | event - ambiguity (slope_2, grassy, foreground, in)
14 | event - ambiguity (several, rams, slope_2, grassy, on)
output: 1 | Are there slopes?
2 | Are there two slopes?
3 | Are there rams?
4 | Is there evergreen?
5 | Is one slope white?
6 | Is one slope grassy?
7 | Is the evergreen measly and underdeveloped?
8 | Is the slope in the background?
9 | Is the slope in the foreground?
10 | Are the several rams on the slope?
11 | Are the several rams grazing on grass?
12 | Is the white slope in the background?
13 | Is the grassy slope in the foreground?
14 | Are the several rams on the grassy slope?

input: A man walks into a room and sits on a chair. A dog follows him.
1 | entity - whole (man)
2 | entity - whole (room)
3 | entity - whole (chair)
4 | entity - whole (dog)
5 | action - (man, walk, room)
6 | action - (man, sit on, chair)
7 | action - (dog, follow, man)
8 | relation - temporal (man, sit, before, walk)
9 | relation - temporal (dog, follows, after, man, sit)
10 | event - temporal (man, walks into a room and sits on a chair, dog follows him)
output: 1 | Is there a man?
2 | Is there a room?
3 | Is there a chair?
4 | Is there a dog?
5 | Does the man walk into the room?
6 | Does the man sit on the chair?
7 | Does the dog follow the man?
8 | Does the man sit before walking?
9 | Does the dog follow after the man sat?
10 | Does the dog follow after the man who walked into a room and sits on a chair?

input: A car is parked by the roadside. Later, it starts moving and drives away.
1 | entity - whole (car)
2 | entity - whole (roadside)
3 | relation - spatial (car, roadside, park)
4 | action - (car, move)
5 | action - (car, drives away)
6 | relation - temporal (car, starts, after, parked)
7 | relation - temporal (car, drive away, after, parked)
8 | event - temporal (car, move, roadside, park, after)
9 | event - temporal (car, drive away, roadside, park, after)
10 | event - temporal (car, starts, parked, move, drive away)
output: 1 | Is there a car?
2 | Is there a roadside?
3 | Does the car park near the roadside?
4 | Des the car move?
5 | Does the car drive away?
6 | Does the car move after being parked?
7 | Does the car drive away after being parked?
8 | Does the car move after being parked near roadside?
9 | Does the car drive away after being parked near roadside?
10 | Is that a same car which parked by the roadsid and then starts moving and drives away?

input: A man is running across a street while carrying a large bag. This is unusual because people typically do not carry large bags while running across streets.
1 | entity - whole (man)
2 | entity - whole (street)
3 | entity - whole (bag)
4 | relation - spatial (man, run, street)
5 | entity - scale (large bag)
6 | relation - spatial (man, carry, large bag)
7 | relation - temporal (man, carry, while, running)
8 | event - ambiguity (man, large bag, carry)
9 | event - temporal (man, run, street, while, carry, large bag)
output: 1 | Is there a man?
2 | Is there a street?
3 | Is there a bag?
4 | Is the man running across the street?
5 | Is a bag large?
6 | Is the man carrying a bag?
7 | Is the man carrying a bag while running?
8 | Is the man carrying a large bag?
9 | Is the man carrying a big bag while running across a street?
\end{lstlisting}
\end{tiny}

\subsection{Dependency Generation}
\begin{tiny}
\begin{lstlisting}
input: A male skateboarder is trying to pull off a trick on the ramp. 
1 | entity - whole (skateboarder)
2 | entity - whole (ramp)
3 | attribute - type (skateboarder, male)
4 | action - (male skateboarder, pull off a trick)
5 | relation - spatial (male skateboarder, ramp, on)
6 | event - ambigity (skateboarder, male, pull off a trick)
7 | event - ambiguity (male skateboarder, ramp, on)
8 | event - ambiguity (skateboarder, pull off a trick, ramp, on)
output: 1 | 0
2 | 0
3 | 1
4 | 1
5 | 1,3
6 | 3,4
7 | 3,5
8 | 4,5

input: A car playing soccer, digital art.
1 | entity - whole (car)
2 | global - (digital art)
3 | action - (car, soccer, play)
output: 1 | 0
2 | 0
3 | 1

input: A set of 2x2 emoji icons with happy, angry, surprised and sobbing faces. The emoji icons look like pigs. All of the pigs are wearing crowns.
1 | entity - whole (emoji icons)
2 | other - count (emoji icons, ==4)
3 | attribute - state (emoji icons, 2x2 grid)
4 | attribute - type (emoji icons, pig)
5 | attribute - state (emoji_1, happy)
6 | attribute - state (emoji_2, angry)
7 | attribute - state (emoji_3, surprised)
8 | attribute - state (emoji_4, sobbing face)
9 | entity - part (pig's crown)
output: 1 | 0
2 | 0
3 | 1
4 | 1
5 | 1
6 | 1
7 | 1
8 | 1
9 | 1,4

input: a photo of bear and dining table; dining table is below bear
1 | global - (photo)
2 | entity - whole (bear)
3 | entity - whole (dining table)
4 | relation - spatial (dining table, bear, below)
output: 1 | 0
2 | 0
3 | 0
4 | 2,3

input: A group of children sitting in the grass with two of them holding a Frisbee .
1 | entity - whole (children)
2 | entity - whole (grass)
3 | entity - whole (frisbee)
4 | attribute - state (children, sit)
5 | relation - spatial (a group of children, grass, sitting in)
6 | entity - part (two of the children)
7 | action - (two of the children, frisbee, hold)
output: 1 | 0
2 | 0
3 | 0
4 | 1
5 | 1,2
6 | 1
7 | 3,6

input: the word 'START' written in chalk on a sidewalk
1 | entity - whole (word)
2 | entity - whole (sidewalk)
3 | other - text rendering (word, "START")
4 | attribute - texture (word, chalk)
5 | relation - spatial (word 'START', sidewalk, on)
output: 1 | 0
2 | 0
3 | 1
4 | 1
5 | 2,3

input: A pear, orange, and two bananas in a wooden bowl.
1 | entity - whole (pear)
2 | entity - whole (orange)
3 | entity - whole (bananas)
4 | other - count (bananas, ==2)
5 | entity - whole (bowl)
6 | attribute - material (bowl, wood)
7 | relation - spatial (pear, bowl, in)
8 | relation - spatial (orange, bowl, in)
9 | relation - spatial (bananas, bowl, in)
10 | relation - spatial (bananas, bowl, in)
11 | event - ambiguity (pear, orange, bananas, ==2, bowl, in)
output: 1 | 0
2 | 0
3 | 0
4 | 0
5 | 0
6 | 0
7 | 1,5
8 | 2,5
9 | 3,5
10 | 4,9
11 | 7,8,10

input: Closeup picture of the front of a clean motorcycle.
1 | entity - whole (motorcycle)
2 | global - (closeup)
3 | global - (picture)
4 | attribute - state (motorcycle, clean)
5 | entity - part (front of the clean motorcycle)
output: 1 | 0
2 | 0
3 | 0
4 | 0
5 | 1

input: a sad man with green hair
1 | entity - whole (man)
2 | entity - part (man's hair)
3 | attribute - state (man, sad)
4 | attribute - color (man's hair, green)
5 | event - ambiguity (man, sad, man's hair, green)
output: 1 | 0
2 | 1
3 | 1
4 | 2
5 | 3,4

input: A commercial airplane with propellers flying through the air.
1 | entity - whole (airplane)
2 | entity - part (airplane's propellers)
3 | action - (airplane, air, fly through)
4 | event - ambiguity (airplane, with propellers, air, fly through)
output: 1 | 0
2 | 0
3 | 1
4 | 2,3

input: A little boy grips a soccer ball in his arms surrounded by other youth soccer players.
1 | entity - whole (boy)
2 | entity - whole (ball)
3 | entity - whole (soccer players)
4 | entity - part (boy's arms)
5 | entity - scale (boy, little)
6 | attribute - type (ball, soccer)
7 | attribute - state (soccer players, youth)
8 | relation - spatial (little boy, ball, grip in his arms)
9 | relation - spatial (little boy gripping the ball in his arms, soccer players, surrounded by)
10 | event - ambiguity (boy's arm, little, ball, soccer, grip in his arms)
11 | event - ambiguity (boy, little, soccer players, youth, surrounded by)
output: 1 | 0
2 | 0
3 | 0
4 | 0
5 | 1
6 | 1
7 | 3
8 | 2,4
9 | 1,3
10 | 4,5,6,8
11 | 5,7,9

input: A traffic light and a signpost at a crossroads intersection near a waterway.
1 | entity - whole (traffic light)
2 | entity - whole (signpost)
3 | entity - whole (crossroads intersection)
4 | entity - whole (waterway)
5 | relation - spatial (traffic light, crossroads intersection, at)
6 | relation - spatial (signpost, crossroads intersection, at)
7 | relation - spatial (traffic light, waterway, near)
8 | relation - spatial (signpost, waterway, near)
9 | relation - spatial (crossroads intersection, waterway, near)
10 | event - ambiguity (traffic light, signpost, crossroads intersection, at)
11 | event - ambiguity (traffic light, crossroads intersection, at, waterway, near)
12 | event - spatial (signpost, crossroads intersection, at, waterway, near)
output: 1 | 0
2 | 0
3 | 0
4 | 0
5 | 1,3
6 | 2,3
7 | 1,4
8 | 2,4
9 | 3,4
10 | 5,6
11 | 5,7
12 | 6,8

input: a photo of dining table and traffic light; traffic light is below dining table
1 | global - (photo)
2 | entity - whole (dining table)
3 | entity - whole (traffic light)
4 | relation - spatial (traffic light, dining table, below)
output: 1 | 0
2 | 0
3 | 0
4 | 2,3

input: A realistic photo of a Pomeranian dressed up like a 1980s professional wrestler with neon green and neon orange face paint and bright green wrestling tights with bright orange boots.
1 | global - (photo)
2 | entity - whole (Pomeranian)
3 | global - (realistic)
4 | entity - part (Pomeranian's costume)
5 | attribute - type (Pomeranian's costume, 1980s professional wrestler)
6 | entity - part (Pomeranian's costume's wrestling tights)
7 | entity - part (Pomeranian's costume's wrestling tights' boots)
8 | entity - part (Pomeranian's facepaint)
9 | attribute - color (Pomeranian's facepaint, neon green)
10 | attribute - color (Pomeranian's facepaint, neon orange)
11 | attribute - color (Pomeranian's costume's wrestling tights, bright green)
12 | attribute - color (Pomeranian's costume's wrestling tights' boots, bright orange)
output: 1 | 0
2 | 0
3 | 0
4 | 2
5 | 4
6 | 4
7 | 4
8 | 2
9 | 8
10 | 8
11 | 6
12 | 7

input: a four-piece band on a stage in front of a small crowd
1 | entity - whole (band)
2 | entity - whole (stage)
3 | entity - whole (crowd)
4 | other - count (band members, ==4)
5 | attribute - shape (crowd, small)
6 | relation - spatial (four-piece band, stage, on)
7 | relation - spatial (four-piece band, crowd, in front of)
8 | relation - spatial (stage, crowd, in front of)
9 | event - ambiguity (band, ==4 picece, stage, on)
10 | event - ambiguity (band, ==4 picece, crowd, small, in front of)
11 | event - ambiguity (stage, crowd, small, in front off)
output: 1 | 0
2 | 0
3 | 0
4 | 1
5 | 3
6 | 2,4
7 | 3,4
8 | 2,3
9 | 2,4
10 | 4,5,7
11 | 2,5,8

input: two laptops, a mouse cord, and a monitor 
1 | entity - whole (laptops)
2 | other - count (laptops, ==2)
3 | entity - whole (mouse coord)
4 | entity - whole (monitor)
output: 1 | 0
2 | 0
3 | 0
4 | 0

input: A red motorcycle parked by paint chipped doors.
1 | entity - whole (motorcycle)
2 | entity - whole (doors)
3 | attribute - color (motorcycle, red)
4 | attribute - state (door, paint chipped)
5 | relation - spatial (red motorcycle, paint chipped door, next to)
6 | attribute - state (motorcycle, parked)
7 | event- ambiguity (motorcycle, red, door, paint chipped, next to)
8 | event- ambiguity (motorcycle, red, parked)
output: 1 | 0
2 | 0
3 | 0
4 | 1
5 | 2
6 | 2,3
7 | 3,4,5
8 | 3,6

input: A cube made of denim. A cube with the texture of denim.
1 | entity - whole (cube)
2 | attribute - material (cube, denim)
3 | attribute - texture (cube, denim)
output: 1 | 0
2 | 1
3 | 1

input: an espresso machine that makes coffee from human souls
1 | entity - whole (espresso machine)
2 | entity - whole (coffee)
3 | entity - whole (human souls)
4 | action - (espresso machine, coffee, make)
5 | attribute - material (coffee, human souls)
6 | event - ambiguity (espresso machine, coffee, make, human souls)
output: 1 | 0
2 | 0
3 | 0
4 | 1,2
5 | 2,3
6 | 4,5

input: Three people standing next to an elephant along a river.
1 | entity - whole (people)
2 | other - count (people, ==3)
3 | entity - whole (elephant)
4 | entity - whole (river)
5 | attribute - state (people, stand)
6 | relation - spatial (three people, elephant, next to)
7 | relation - spatial (people, river, next to)
8 | relation - spatial (elephant, river, next to)
9 | event - ambiguity (people, ==3, stand)
10 | event - ambiguity (people, ==3, elephant, next to)
11 | event - ambiguity (people, ==3, river, next to)
12 | event - ambiguity (people, stand, elephant, next to)
13 | event - ambiguity (people, stand, river, next to)
14 | event - ambiguity (people, elephant, next to, river, next to)
output: 1 | 0
2 | 1
3 | 0
4 | 0
5 | 1
6 | 1,3
7 | 1,4
8 | 2,4
9 | 2,5
10 | 2,6
11 | 2,7
12 | 5,6
13 | 5,7
14 | 6,7

input: Aerial view of downtown Manhattan, but with Millennium Wheel next to the Statue of Liberty. The Great Pyramid is on a sandy island near the buildings.
1 | entity - (downtown Manhattan)
2 | entity - (Millennium Wheel)
3 | entity - (the Statue of the Liberty)
4 | entity - (the Great Pyramid)
5 | entity - (island)
6 | entity - (buildings)
7 | global - (aerial view)
8 | attribute - texture (island, sandy)
9 | relation - spatial (Millennium Wheel, the Statue of Liberty, next to)
10 | relation - spatial (the Great Pyramid, island, on)
11 | relation - spatial (the Great Pyramid, buildings, near)
12 | event - ambiguity (the Great Pyramid, island, on, buildings, near)
output: 1 | 0
2 | 0
3 | 0
4 | 0
5 | 0
6 | 0
7 | 0
8 | 5
9 | 2,3
10 | 4,5
11 | 4,6
12 | 10,11

input: A laptop with external keyboard, mouse, phone and photo on a desk.
1 | entity - whole (laptop)
2 | entity - whole (keyboard)
3 | entity - whole (mouse)
4 | entity - whole (phone)
5 | entity - whole (photo)
6 | entity - whole (desk)
7 | attribute - type (keyboard, external)
8 | relation - spatial (laptop, desk, on)
9 | relation - spatial (keyboard, desk, on)
10 | relation - spatial (mouse, desk, on)
11 | relation - spatial (phone, desk, on)
12 | relation - spatial (photo, desk, on)
13 | event - ambiguity (laptop, external keyboard, mouse, phone, photo, desk, on)
output: 1 | 0
2 | 0
3 | 0
4 | 0
5 | 0
6 | 0
7 | 0
8 | 1,6
9 | 2,6
10 | 3,6
11 | 4,6
12 | 5,6
13 | 8,9,10,11,12

input: A white slope covers the background, while the foreground features a grassy slope with several rams grazing and one measly and underdeveloped evergreen in the foreground.  
1 | entity - whole (slopes)
2 | other - count (slopes, ==2)
3 | entity - whole (rams)
4 | entity - whole (evergreen)
5 | attribute - color (slope_1, white)
6 | attribute - texture (slope_2, grassy)
7 | attribute - state (evergreen, measly and underdeveloped)
8 | relation - spatial (slope_1, background, in)
9 | relation - spatial (slope_2, foreground, in)
10 | relation - spatial (several, rams, grassy slope_2, on)
11 | attribute - state (several rams, graze)
12 | event - ambiguity (slope_1, white, background, in)
13 | event - ambiguity (slope_2, grassy, foreground, in)
14 | event - ambiguity (several, rams, slope_2, grassy, on)
output: 1 | 0
2 | 1
3 | 0
4 | 0
5 | 1
6 | 1
7 | 4
8 | 5
9 | 1
10 | 1
11 | 1,3
12 | 5,8
13 | 6,9
14 | 6,10

input: A man walks into a room and sits on a chair. A dog follows him.
1 | entity - whole (man)
2 | entity - whole (room)
3 | entity - whole (chair)
4 | entity - whole (dog)
5 | action - (man, walk, room)
6 | action - (man, sit on, chair)
7 | action - (dog, follow, man)
8 | relation - temporal (man, sit, before, walk)
9 | relation - temporal (dog, follows, after, man, sit)
10 | event - temporal (man, walks into a room and sits on a chair, dog follows him)
output: 1 | 0
2 | 0
3 | 0
4 | 0
5 | 1,2
6 | 1,3
7 | 1,4
8 | 5,7
9 | 6,7
10 | 8,9

input: A car is parked by the roadside. Later, it starts moving and drives away.
1 | entity - whole (car)
2 | entity - whole (roadside)
3 | relation - spatial (car, roadside, park)
4 | action - (car, move)
5 | action - (car, drives away)
6 | relation - temporal (car, starts, after, parked)
7 | relation - temporal (car, drive away, after, parked)
8 | event - temporal (car, move, roadside, park, after)
9 | event - temporal (car, drive away, roadside, park, after)
10 | event - temporal (car, starts, parked, move, drive away)
output: 1 | 0
2 | 0
3 | 1,2
4 | 1
5 | 1
6 | 1,4
7 | 1, 5
8 | 3,6
9 | 3,7
10 | 6,7

input: A man is running across a street while carrying a large bag. This is unusual because people typically do not carry large bags while running across streets.
1 | entity - whole (man)
2 | entity - whole (street)
3 | entity - whole (bag)
4 | relation - spatial (man, run, street)
5 | entity - scale (large bag)
6 | relation - spatial (man, carry, large bag)
7 | relation - temporal (man, carry, while, running)
8 | event - ambiguity (man, large bag, carry)
9 | event - temporal (man, run, street, while, carry, large bag)
output: 1 | 0
2 | 0
3 | 0
4 | 1, 2
5 | 3
6 | 1, 5
7 | 4, 7
8 | 5,6
9 | 4, 7
\end{lstlisting}
\end{tiny}